\documentclass{article} 
\usepackage{iclr2027_conference,times}

\usepackage{amsmath,amsfonts,bm}

\def\eqref#1{equation~\ref{#1}}

\def\1{\bm{1}}

\DeclareMathAlphabet{\mathsfit}{\encodingdefault}{\sfdefault}{m}{sl}
\SetMathAlphabet{\mathsfit}{bold}{\encodingdefault}{\sfdefault}{bx}{n}

\usepackage{hyperref}
\usepackage{url}

\usepackage[utf8]{inputenc}
\usepackage[T1]{fontenc}
\usepackage{booktabs}
\usepackage{array}
\usepackage{amsfonts}
\usepackage{amsmath}
\usepackage{nicefrac}
\usepackage{microtype}
\usepackage{xcolor}
\usepackage{graphicx}
\usepackage{multirow}
\usepackage{placeins}
\usepackage{soul}

\newif\ifmarkupcolors
\markupcolorsfalse
\ifmarkupcolors
\newcommand\todo[1]{{\color{red}{#1}}}%
\newcommand\ok[1]{{\color{blue}{#1}}}%
\newcommand\rev[1]{{\color{orange}{#1}}}%
\newcommand\notouch[1]{{\color{blue}{#1}}}%
\else
\newcommand\todo[1]{#1}%
\newcommand\ok[1]{#1}%
\newcommand\rev[1]{#1}%
\newcommand\notouch[1]{#1}%
\fi

\IfFileExists{tables/key_numbers.tex}{
\newcommand{\knAlphaCreativity}{0.25}
\newcommand{\knIccCkCreativity}{0.80}

\newcommand{\knAlphaRecognizability}{0.40}

\newcommand{\knAlphaAudraCorpus}{0.40}

\newcommand{\knConstantRatersCreativity}{3}

\newcommand{\knBestModel}{GPT-6 Astra}
\newcommand{\knBestModelMean}{0.81}
\newcommand{\knBestModelCI}{$[0.80, 0.83]$}
\newcommand{\knWorstModel}{Gemini 3.5 Flash Lite}
\newcommand{\knWorstModelMean}{0.35}
\newcommand{\knWorstModelCI}{$[0.33, 0.37]$}

\newcommand{\knNHumanDrawings}{300}
\newcommand{\knHumanMean}{0.41}
\newcommand{\knAgentMean}{0.56}
\newcommand{\knAgentMinusHuman}{$+0.15$}
\newcommand{\knAgentMinusHumanCI}{$[0.13, 0.17]$}
\newcommand{\knAgentHumanD}{0.88}
\newcommand{\knAgentPercentileOfHuman}{83rd}

\newcommand{\knHumanMeanRecog}{0.53}
\newcommand{\knAgentMeanRecog}{0.45}
\newcommand{\knAgentMinusHumanRecog}{$-0.08$}
\newcommand{\knAgentMinusHumanRecogCI}{$[-0.11, -0.06]$}
\newcommand{\knAgentHumanDRecog}{-0.41}
\newcommand{\knAgentPercentileOfHumanRecog}{36th}

\newcommand{\knCreatRecogR}{0.61}
\newcommand{\knCreatRecogCI}{$[0.57, 0.64]$}

\newcommand{\knCreatRecogWithinR}{0.38}
\newcommand{\knCreatRecogWithinCI}{$[0.34, 0.42]$}

\newcommand{\knModelEffectChiSq}{3096.9}
\newcommand{\knModelEffectDf}{13}
\newcommand{\knModelEffectP}{< 0.001}

\newcommand{\knVidraTestR}{0.86}

\newcommand{\knVidraTestRSd}{0.01}
\newcommand{\knAudraInkRhoAgent}{0.86}
\newcommand{\knVidraInkRhoAgent}{0.77}
\newcommand{\knAudraInkRhoHuman}{0.69}

\newcommand{\knCreativityInkR}{0.40}

\newcommand{\knRecognizabilityInkR}{0.10}

\newcommand{\knHumanRatingInkN}{300}
\newcommand{\knHumanCreativityInkR}{0.37}

\newcommand{\knHumanRecognizabilityInkR}{0.13}

\newcommand{\knAdaptModelR}{0.83}

\newcommand{\knAdaptModelInkRho}{0.70}
\newcommand{\knAdaptTaskR}{0.84}

\newcommand{\knAdaptModelRankRho}{0.99}

\newcommand{\knAdaptedTestR}{0.85}

\newcommand{\knPPCallsPooledRho}{0.06}

\newcommand{\knPPCallsWithinRho}{-0.01}

\newcommand{\knPPOpsWithinRho}{0.06}
\newcommand{\knPPOpsWithinCI}{$[0.02, 0.11]$}
\newcommand{\knPPOpsPerRoundPooledRho}{0.59}
\newcommand{\knPPOpsPerRoundPooledCI}{$[0.56, 0.62]$}

\newcommand{\knUndoTrials}{480}
\newcommand{\knEraseTrials}{185}
\newcommand{\knToolErrorTrials}{11}
\newcommand{\knMalformedTrials}{87}
\newcommand{\knUndoTrialsPct}{22.9\%}
\newcommand{\knEraseTrialsPct}{8.8\%}

\newcommand{\knUndoActiveMin}{48}
\newcommand{\knUndoTopModel}{Muse Spark 1.3}
\newcommand{\knUndoTopModelUndo}{105}
\newcommand{\knUndoTopModelErase}{75}

\newcommand{\knUndoOtherMax}{3}

\newcommand{\knErrorTopModel}{Qwen3.5-9B}
\newcommand{\knErrorTopModelToolErrors}{6}
\newcommand{\knErrorTopModelMalformed}{58}

}{}
\IfFileExists{tables/ablation_briefs.tex}{
\newcommand{\ablSvgCondMean}{0.68}
\newcommand{\ablSvgBaseMean}{0.61}
\newcommand{\ablSvgDelta}{$+0.07$}
\newcommand{\ablSvgCI}{$[0.04, 0.10]$}
\newcommand{\ablSvgDeltaSd}{$0.43$}

\newcommand{\ablSvgRecogCondMean}{0.40}
\newcommand{\ablSvgRecogBaseMean}{0.33}
\newcommand{\ablSvgRecogDelta}{$+0.07$}
\newcommand{\ablSvgRecogCI}{$[0.04, 0.11]$}
\newcommand{\ablSvgRecogDeltaSd}{$0.36$}
\newcommand{\ablSvgRecogN}{30}
\newcommand{\ablOracleCondMean}{0.51}
\newcommand{\ablOracleBaseMean}{0.33}
\newcommand{\ablOracleDelta}{$+0.18$}
\newcommand{\ablOracleCI}{$[0.12, 0.24]$}
\newcommand{\ablOracleDeltaSd}{$0.86$}

\newcommand{\ablBlankCondMean}{0.59}
\newcommand{\ablBlankBaseMean}{0.61}
\newcommand{\ablBlankDelta}{$-0.02$}
\newcommand{\ablBlankCI}{$[-0.06, 0.03]$}
\newcommand{\ablBlankDeltaSd}{$-0.12$}

\newcommand{\ablFramingDepictiveCondMean}{0.52}
\newcommand{\ablFramingDepictiveBaseMean}{0.35}
\newcommand{\ablFramingDepictiveDelta}{$+0.17$}
\newcommand{\ablFramingDepictiveCI}{$[0.10, 0.23]$}
\newcommand{\ablFramingDepictiveDeltaSd}{$0.83$}

\newcommand{\ablFramingExampleCondMean}{0.45}

\newcommand{\ablFramingExampleDelta}{$+0.10$}
\newcommand{\ablFramingExampleCI}{$[0.02, 0.17]$}
\newcommand{\ablFramingExampleDeltaSd}{$0.50$}

\newcommand{\ablPrimitivesCondMean}{0.51}
\newcommand{\ablPrimitivesBaseMean}{0.59}
\newcommand{\ablPrimitivesDelta}{$-0.08$}
\newcommand{\ablPrimitivesCI}{$[-0.14, -0.03]$}
\newcommand{\ablPrimitivesDeltaSd}{$-0.48$}

}{}
\providecommand{\knAdaptModelR}{--}
\providecommand{\knAdaptTaskR}{--}
\providecommand{\knAdaptModelInkRho}{--}

\providecommand{\knConstantRatersCreativity}{--}

\providecommand{\knVidraTestR}{--}
\providecommand{\knVidraTestRSd}{--}
\providecommand{\knVidraInkRhoAgent}{--}
\providecommand{\knAudraInkRhoAgent}{--}
\providecommand{\knAudraInkRhoHuman}{--}
\providecommand{\knAdaptModelRankRho}{--}
\providecommand{\knHumanRatingInkN}{--}
\providecommand{\knHumanCreativityInkR}{--}

\providecommand{\knHumanRecognizabilityInkR}{--}

\providecommand{\knModelEffectChiSq}{--}
\providecommand{\knModelEffectDf}{--}
\providecommand{\knModelEffectP}{--}

\providecommand{\knCreativityInkR}{--}

\providecommand{\knRecognizabilityInkR}{--}

\providecommand{\knIccCkCreativity}{--}

\providecommand{\knAlphaAudraCorpus}{--}
\providecommand{\knBestModel}{--}
\providecommand{\knBestModelMean}{--}
\providecommand{\knBestModelCI}{--}
\providecommand{\knWorstModel}{--}
\providecommand{\knWorstModelMean}{--}
\providecommand{\knWorstModelCI}{--}

\providecommand{\ablSvgDelta}{--}
\providecommand{\ablSvgCI}{--}
\providecommand{\ablSvgDeltaSd}{--}

\providecommand{\ablPrimitivesDelta}{--}
\providecommand{\ablPrimitivesCI}{--}
\providecommand{\ablPrimitivesDeltaSd}{--}

\providecommand{\ablBlankDelta}{--}
\providecommand{\ablBlankCI}{--}
\providecommand{\ablBlankDeltaSd}{--}

\providecommand{\ablOracleDelta}{--}
\providecommand{\ablOracleCI}{--}
\providecommand{\ablOracleDeltaSd}{--}

\providecommand{\ablFramingDepictiveDelta}{--}
\providecommand{\ablFramingDepictiveCI}{--}

\providecommand{\ablFramingExampleDelta}{--}
\providecommand{\ablFramingExampleCI}{--}

\newcommand{\modelname}[1]{#1}

\newcommand{\mdlFable}{claude-fable-5}
\newcommand{\mdlOpus}{claude-opus-5}
\newcommand{\mdlSonnet}{claude-sonnet-5}

\title{PainterBench: A Figural Divergent-Thinking \\ Benchmark for Tool-Using Language Models 
}

\author{Shane K.A. Dalumura Hettige \\
Computer Science and Engineering \\
University of Oulu \\
Oulu, Finland \\
\texttt{shane.dalumurahettige@student.oulu.fi} \\
\And
Jonas Oppenlaender \\
Centre for Applied Computing \\
University of Oulu \\
Oulu, Finland \\
\texttt{jonas.oppenlaender@oulu.fi}
}

\iclrfinalcopy

\begin{document}
\maketitle
\lhead{Preprint}

\begin{abstract}
	Figural divergent thinking is the ability to develop a given shape fragment into an original drawing.
	In humans, this ability is assessed with \textit{incomplete-drawing tasks}.
	We introduce \textbf{PainterBench}, a benchmark that ports the incomplete-drawing task to the agentic setting.
	The agent draws on a canvas through tool calls and observes the result after every turn.
	The canvas includes a starting shape which cannot be erased, and the agent's goal is to incorporate this shape into the most original drawing it can produce.
	The task is open-ended, and 
	the agent itself decides when the drawing is finished.
	The benchmark tests incremental visual planning over a short horizon and the transfer of creative ability from pretraining to multi-turn tool use.
	We evaluate 14 multimodal language models from small to frontier scale.
	Across the primary study and six sensitivity analyses, we collect 2{,}700 drawings and crowdsource creativity and recognizability ratings for every drawing and for 300 human reference drawings.
	We also present \textbf{ViDrA-adapted}, an automated scorer that predicts human creativity ratings of agent drawings ($r = \knAdaptedTestR$ on random held-out test split). 
	Figural divergent thinking varies widely across the 14 models, and GPT-6 Astra produces the most creative drawings.
	Relative to the human drawings, the agent drawings score higher in creativity but lower in recognizability.
	We release the final drawings, 
	per-round canvas snapshots, 
	tool call traces, stimulus bank, 
	benchmark harness,
	crowdsourced ratings (N~=~72,000), and ViDrA checkpoint.
\end{abstract}

\section{Introduction}%
\label{sec:introduction}%
%
\notouch{%
	An early demonstration of creative behavior in large language models was the prompt ``draw a unicorn in TikZ'' \citep{bubeck2023sparksartificialgeneralintelligence}.
	Bubeck et al. presented the resulting 
	figure as evidence of the model's understanding of visual and geometric concepts,
	despite text-only training.
	The demonstration, however, was one zero-shot prompt, and it tested the model's ability to output text, the modality it was trained on.
	We argue evaluating a model's 
	creative ability requires testing beyond the reproduction of patterns acquired during pretraining.
	Further, today's models are deployed as tool-calling agents that observe the result of each action and revise their work over many turns.
	Tool calls can present the model with tasks it has not encountered during pretraining.
	Whether pretrained creative ability carries over to the tool-calling agentic setting has not been tested.%
}%

\ok{%
	Prior studies evaluate creativity in language models in the textual domain with established divergent thinking tests, such as Guilford's Alternative Uses Task \citep{guilford1967nature,patterson2024audra,stevenson2022puttinggpt3screativityalternative,haase2026withinmodelvsbetweenpromptvariability,10.1145/3772363.3799284,schapiro2026creativityneuro,brainwriting}.
	However, there are known limitations to this divergent thinking test, such as scores depending on verbal fluency and limited predictive validity \citep{Zeng04022011,barbot2018mtci}, and
	divergent thinking is more than just text production.
	The closest prior work, SketchAgent~\citep{vinker2025sketchagent}, prompts a multimodal language model to draw through a bespoke sketching language.
	The system is created for iterative conversational refinement of sketches,
	emits the full stroke sequence in string-based actions, and is evaluated on recognizability rather than creativity.%
}%

\notouch{%
	Human creativity research provides established instruments for assessing whether a drawing is creative.
	Figural divergent thinking has been assessed for decades with \textit{incomplete-drawing tasks}, from the Torrance 
	test battery~\citep{TorranceTest} to the Multi-Trial Creative Ideation (MTCI) task~\citep{barbot2018mtci}. In the latter, a person turns a given shape fragment into the most original drawing they can think of.
	This task also comes with a validated automated scorer, AuDrA,
	which predicts human creativity ratings of drawings 
	\citep{patterson2024audra}.%
}%

\ok{%
	We introduce \textbf{PainterBench}, a benchmark harness that evaluates tool-using agents on the incomplete-drawing task.
	The agent draws via discrete tool calls over a library of drawing operations, observes a rendering of the canvas after every turn, and itself declares the drawing finished.
	Each trial 
	pre-seeds the canvas with one of 30 stimuli (starting
	shapes; see \autoref{fig:stimuli}) that cannot be erased.
	The novel drawing tools, not SVG or TikZ markup, are the agent's 
	medium, so performance cannot follow from a representation 
	encountered during pretraining.
	The benchmark, therefore, tests whether the model's creative ability acquired in pretraining survives the transfer to acting through tools over many turns.
	Because the agent declares its own drawing finished, each drawing is a product the model judged complete.
	The tool call trace 
	makes the model's creative process observable.%
}%

\ok{%
	Across 14 multimodal language models, we find that creative ability survives the transfer in part.
	Under the rating protocol of the human task, the agent drawings score above a reference sample of human drawings on creativity but below it on recognizability.
	The tool-call traces show the same imbalance in the creative process.
	The agents show little revision, and each model returns to a small set of ideas across its independent trials.
	The agent drawings lie far from the distribution of human drawings, and automated creativity scorers trained on human drawings correlate with ink-on-canvas on agent drawings.
}%

\ok{The contributions of this paper are as follows:}%
\begin{itemize}%
	\item \ok{%
		We present \textbf{PainterBench}, a benchmark that ports the 
		task of figural divergent-thinking assessment \citep{barbot2018mtci} to the agentic setting.
		We release a bank of 30 stimuli in MTCI's two task types, and
		we define tool-call process markers that make the creative process measurable.
		We release a dataset that includes 2{,}100 agent drawings with crowdsourced creativity and recognizability ratings, full tool call traces, 
		and \ok{29{,}708} per-round snapshots. 
	}%
	
	\item
	\ok{%
		We release \textbf{ViDrA-adapted}, an 
		automated creativity scorer, fit to the public AuDrA corpus  of 
		human drawings \citep{patterson2024audra} and adapted to agent drawings.
		The adapted scorer reaches $r = \knAdaptModelR$ under leave-one-model-out and $r = \knAdaptTaskR$ under leave-one-stimulus-out cross-validation on agent drawings.
	}%
	
	\item
	\ok{%
		Using PainterBench and ViDrA, we evaluate \textbf{figural divergent thinking in 14 
			multimodal language models} from small to frontier scale.
		This evaluation measures whether pretrained creative ability transfers to the agentic setting, in which the agent plans the drawing incrementally over a short visual horizon.
		In six sensitivity analyses, we test how the results depend on the harness design choices.
		Finally, we discuss qualitative findings and frequently occurring failure modes in agentic drawings.
	}%
\end{itemize}

\section{Related Work}%
\label{sec:relatedwork}%
%
\textbf{Machine drawing systems.}
\ok{%
	Autonomous drawing systems, from hand-crafted procedural rules~\citep{10.5555/2887965.2888115} to learned stroke-based painters and sketch models~\citep{ganin2018synthesizing,huang2019learning,ha2018neural}, were trained specifically to draw and do not follow novel instructions zero-shot.
	A parallel line of research asks whether pretrained large language models (LLMs) can produce visual output directly, either as graphics code or as drawing actions~\citep{belouadi2024automatikz,belouadi2024detikzify,zou2024vgbench,cai2024,zini2025svgauge,sharma2024visioncheckup}.
	Recent LLMs reach human-level visualization literacy but violate instructions and graphical integrity~\citep{seto2026llmsvisualizationliteracy}, and perception remains a source of systematic error in spatial and compositional tasks~\citep{lu2026visualprimitives,park2024designing}.
	Within this line, SketchAgent~\citep{vinker2025sketchagent} prompts a multimodal model to emit stroke coordinates rendered as B\'ezier curves. LTD-Bench~\citep{lin2025ltdbench}, TurtleBench~\citep{rismanchian2025turtlebench}, and DrawingBench~\citep{kim2026drawingbench} score drawings produced as dot matrices, Turtle programs, and GUI mouse actions, respectively.
	3DrawAgent~\citep{xiao20263drawagent} extends the setting to 3D curves.
	These studies score fidelity, recognizability, or spatial accuracy against a reference, not creativity, and none exposes the model's native function-calling interface with per-turn visual feedback.%
}%

\textbf{Figural creativity assessment.}
\ok{%
	Divergent thinking has been the standard behavioral measure of creative ideation
	in research since \citet{guilford1950creativity}.
	In figural divergent thinking, the standard tests are incomplete-drawing tasks, such as the Torrance figural tests~\citep{TorranceTest,kim2006trust},
	the Test for Creative Thinking-Drawing Production (TCT-DP) \citep{TCTDP,urban2005tctdp},
	and the ``Multi-Trial Creative Ideation'' assessment framework (MTCI)~\citep{barbot2018mtci}.
	These tests measure creativity through drawing production rather than verbal responses.
}%
\ok{%
	MTCI evaluates figural divergent thinking by presenting one stimulus per trial, collecting a single self-paced drawing from participants, and segmenting the drawing process into three time-based phases (exploration, production, and verification). 
	Responses are scored by human judges under consensual-assessment-style, in which untrained judges rate creativity by their own subjective standard protocols \citep{amabile1982social,silvia2008assessing}.
	AuDrA~\citep{patterson2024audra} applies the MTCI protocol at scale, collecting creativity ratings from 
	human raters on a five-point scale for a public corpus of over 13{,}000 MTCI drawings.
	PainterBench adopts the MTCI trial structure, the participant instruction, and the rating protocol of \citet{patterson2024audra}.
}%

\textbf{Automated creativity scoring.}
\ok{%
	Automated scoring emerged first in the verbal domain, where semantic-distance systems such as SemDis predict human originality ratings~\citep{beaty2021semdis}.
	In the figural domain, \citet{cropley2025automated} classified TCT-DP drawings with a convolutional network.
	Closest to our setting, \citet{nath2025pencils} compare drawings by children, adults, and AI on MTCI stimuli, with the AI drawings produced by one-shot generation rather than by sequential tool use.
	\citet{acar2025automated} score Torrance-figural and MTCI drawings with vision transformers, and AuDrA~\citep{patterson2024audra} predicts human creativity ratings for the drawings and maps a drawing to a continuous creativity score, normalized to $[0,1]$.
	We train our own scorer, ViDrA, on the public AuDrA corpus and validate both ViDrA and AuDrA on agent-generated drawings, a domain neither has been tested on.%
}%

\textbf{Creativity benchmarks for language models.}
\ok{%
	Creativity evaluations for language models 
	often adapt established tests such as the Divergent Association Task (DAT)~\citep{olson2021naming} or the Alternative Uses Task (AUT)~\citep{guilford1967nature}.
	\citet{chakrabarty2024artifice} apply a rating protocol derived from the Torrance tests to short stories, and language-model stories pass far fewer of its tests than stories by professional writers.
	NeoCoder~\citep{lu2025neocoder} elicits creative programs by imposing successive constraints that deny the model its previous solution, and scores the responses against a reference set of human solutions.
	These tests are text-based and do not assess creative composition in a visual medium.
	CreativityBench~\citep{qian2026creativitybench} asks a model to repurpose an object by reasoning about its affordances rather than its canonical use.
	Both benchmarks, like benchmarks of function calling itself~\citep{qin2023toolllm,NEURIPS2024_e4c61f57,patil2025the}, pose tasks that admit a correct answer, which is what allows automatic scoring.
	Figural divergent thinking, however, is open-ended and admits no correct answer.
	Reference-free metrics for open-ended text score originality by attributing machine text to web text~\citep{lu2025salieri}, but such n-gram novelty diverges from expert judgments of creativity~\citep{saakyan2026novelty}.
	Responses must therefore be rated, and PainterBench adopts the rating protocol of an established human assessment.%
}%
%
%
\section{PainterBench: A Figural Divergent-Thinking Benchmark}%
\begin{figure*}[!htb]
	\centering
	\includegraphics[width=.78\textwidth]{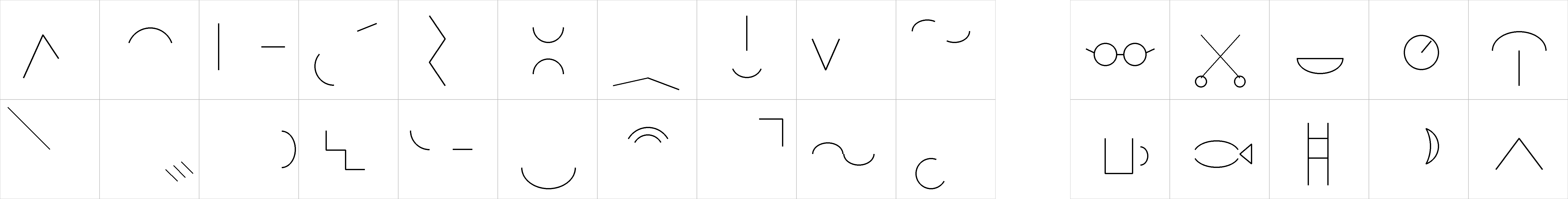}
	\caption{PainterBench stimuli: 
		20 incomplete shapes (left) and 10 object transformations (right).
	}
	\label{fig:stimuli}%
\end{figure*}

\label{sec:painterbench}%
%
\label{sec:task}%
\textbf{Task.}
\ok{%
	PainterBench ports the MTCI drawing task~\citep{barbot2018mtci,patterson2024audra} to tool-using agents.
	The agent draws on a 400$\times$400 pixel canvas, annotated with pixel coordinates, through a sequence of tool calls over drawing, erase, and undo operations (see \autoref{tab:tools}).
	The canvas size, stroke width, and the black-on-white medium reproduce the drawing geometry of the AuDrA corpus \citep{patterson2024audra}.
	%
	%
	A \emph{trial} is one run of the agent loop on one stimulus and produces one drawing.
	A \emph{replicate} is an independent trial on the same stimulus with identical inputs.
	Each of the 30 trials presents the agent with a canvas containing one starting stimulus (see \autoref{fig:stimuli}).
	The agent is instructed to create the most original drawing that incorporates the given shape.
	%
	%
	A \textit{round} is one model turn, after which the model inspects the canvas.
	As in MTCI, the task is open-ended. The agent ends the trial by calling the \texttt{drawing\_finished} tool.%
}%

\begin{table}[bht]
	\caption{The 14 tools available to the drawing agent. Coordinates are in pixel units on the canvas.
	}%
	\label{tab:tools}
	\centering
	\setlength{\tabcolsep}{4pt}
	\resizebox{\linewidth}{!}{%
		\begin{tabular}{llll}
			\toprule
			Tool                        & Description                        & Tool                        & Description \\
			\midrule
			\texttt{draw\_dots}         & Place dots at specific coordinates                 & \texttt{draw\_rounded\_rectangles} & Rectangles with a corner radius \\
			\texttt{draw\_lines}        & Segments between two points            & \texttt{draw\_circles}      & From a center and a radius \\
			\texttt{draw\_polylines}    & Connected stroke through points    & \texttt{draw\_ellipses}     & From a bounding box \\
			\texttt{draw\_polygons}     & Closed shape through points        & \texttt{draw\_arcs}         & From a box and angle range \\
			\texttt{draw\_regular\_polygons} & From center, radius, sides, rotation & \texttt{draw\_pieslices}    & Arc closed to the center \\
			\texttt{draw\_rectangles}   & From two opposite corners          & \texttt{draw\_chords}       & Arc closed by a straight line \\
			\midrule
			\texttt{undo\_last\_action} & Revert the last call           & \texttt{drawing\_finished}  & Record title, end trial \\
			\bottomrule
		\end{tabular}
	}%
\end{table}

\label{sec:harness}%


\ok{%
	Each round, the agent receives the following context:
	1) the drawing task in the system prompt (see Appendix \ref{app:prompts}),
	2) an instruction in the user message,
	3) the current canvas,
	4) a history of ten most recent tool calls and their results, and
	5) a visual history of the three most recent canvases, stitched together into one image.
	By design, we do not ask the model to create a written plan.
	This 
	frames the task as \textbf{incremental visual planning over a short horizon}.
}%

\textbf{Drawing harness and tools.}
\ok{%
	The agent has access to 14 tools (Table~\ref{tab:tools}).
	Each tool accepts one operation or a batch of operations of its type (full call signatures in Appendix~\ref{app:tools}).
	A failed call is reported to the agent in the next round.
	%
	All strokes are black at a fixed width of 5 pixels. 
	%
	%
	%
	Features of general drawing programs, such as text rendering, layers, geometric transformations, brushes, variable stroke width, and flood fill, are excluded to match the drawing procedures of MTCI~\citep{barbot2018mtci} and the AuDrA corpus \citep{patterson2024audra}.
	Every drawing tool takes an optional \texttt{erase} flag 
	which lays white along the path the shape would have drawn. 
	The stimulus is re-stamped after every canvas operation, so it cannot be erased.
	The meta-tool ending the trial collects a title for the drawing, following the MTCI protocol \citep{barbot2018mtci}. We analyze the titles in Appendix~\ref{app:titles}.
	The trial continues until the model calls 
	\texttt{drawing\_finished}, or until no further tool calls are made.
}%

\textbf{Stimulus bank.}
\label{sec:tasks}
\ok{%
	The stimulus bank consists of 30 starting shapes (see Figure~\ref{fig:stimuli}) in MTCI's two task types.
	The 20 \emph{incomplete shapes} (is01--is20) are 
	stroke fragments
	that do not form a closed object.
	The ten \emph{object-transformation} items (ot01--ot10) are contour outlines of recognizable objects (e.g., glasses, scissors).
	The stimuli were procedurally generated following MTCI's design principles \citep{barbot2018mtci}, and no item reproduces an MTCI item exactly.
}%

\vspace{-.5\baselineskip}%
\section{Experiments}%
\label{sec:eval}%

\label{sec:setup}%

\ok{%
	We evaluate 14~multimodal language models (see Appendix~\ref{app:models}) spanning seven providers and a range of capability tiers.
	Each model completes all 30 stimuli five times, for a total of 2{,}100 trials,
	with 
	five runs per stimulus, differing only through sampling
}%

\textbf{Crowdsourced creativity ratings.} 
\ok{%
	We collect ratings for the main study's 2{,}100 agent drawings, the sensitivity analysis (600 drawings), and 300 human reference drawings sampled from parts of the AuDrA corpus outside our own scorer's training data
	(200 from the primary set's held-out test split and 100 from AuDrA's far-generalization set).
	Ratings are collected on CloudResearch 
	\citep{litman2017turkprime}, a crowdsourcing platform.
	The rater pool is gender-balanced, and workers are required to have completed 100 prior tasks with an acceptance rate of 95\%.
	Following \citet{patterson2024audra}, we assess inter-rater reliability (agreement among annotators), measured  
	with the intraclass correlation 
	ICC(C,k)~\citep{koo2016guideline}, and report Krippendorff's ordinal $\alpha$~\citep{krippendorff2011computing}, an agreement coefficient for ordered ratings, where $\alpha = 1$ is perfect agreement and $\alpha = 0$ is agreement expected by chance.
	A pilot determines $k = 12$ raters per drawing (see Appendix~\ref{app:power}).
	Each rater rates one batch of 30 drawings on two questions, which amounts to $72{,}000$ crowdsourced ratings in total.%
}%

\ok{%
	We collect ratings under the AuDrA rater protocol \citep{patterson2024audra}, which follows consensual-assessment-style subjective scoring~\citep{amabile1982social,silvia2008assessing}.
	First, each drawing is rated on creativity on a five-point scale (``How creative is this drawing?'', from 1 -- \emph{Not At All Creative}, to 5 -- \emph{Very Creative} \citep{patterson2024audra,10.3389/fpsyg.2019.00985}.
	Raters are 
	instructed to rate the creativity of the idea expressed in the drawing, not the technical proficiency of the drawing (see Appendix~\ref{app:raterinstruction}).
	Second, raters assess recognizability (``Does this drawing show a recognizable object or scene?'', from 1 -- \emph{Not At All Recognizable} to 5 -- \emph{Very Recognizable}).
	Raters are 
	not told which drawings are machine-generated,
	because people may be biased against AI-generated content \citep{10.1037/aca0000136,10.1145/3334480.3382892}.
	Unlike in AuDrA, raters are not shown the drawing titles.
	We exclude raters who gave every drawing the same rating (N~=~\knConstantRatersCreativity{}).
	Following \citet{patterson2024audra}, each rater's ratings are z-scored across all their responses 
	to adjust for differences in how raters use the scale
	\citep{LONG201513}, averaged per drawing, and min--max normalized to $[0,1]$.
	For the creativity question, we call this \emph{rated creativity}.
	The term \emph{creativity score} refers to what the automated scorer predicts.
}%

\ok{%
	Ordinal $\alpha$ among individual raters is \knAlphaCreativity{} for creativity, against $\alpha = \knAlphaAudraCorpus$ in the AuDrA rater pool \citep{patterson2024audra}, and \knAlphaRecognizability{} for recognizability.
	The 
	12 ratings per drawing (the composite),
	on which all analyses are based, reaches ICC(C,k) $= \knIccCkCreativity$ (see Appendix~\ref{app:power}). The residual rating noise attenuates correlations and widens confidence intervals, and does not bias the per-model means.
}%

\textbf{Process measures.} 
\label{sec:process}%
%
\ok{%
	MTCI's key process measure is response time
	\citep{barbot2018mtci}.
	For agents, wall-clock time confounds ideation with inference latency and provider load.
	Instead, we define eleven process markers (Table~\ref{tab:markerdefs} in Appendix~\ref{app:processproduct}), and report elapsed time 
	for reference only.
}%
\ok{%
	For model comparison, we combine four markers (mean tool calls, mean drawing operations, mean rounds to completion, and mean tool diversity) into one number per trial.
	We call this the \emph{effort index}, a measure of the amount of drawing activity in a trial.
	With standardized components and no criterion, we weight the four markers uniformly 
	\citep{dawes1979robust}.
	The effort index of trial $i$ is $\mathrm{effort}_i = |M|^{-1} \sum_{m \in M} (m_i - \bar{m})/s_m$,
	where $M$ is the set of four included markers, $m_i$ is the value of marker $m$ on trial $i$, and $\bar{m}$ and $s_m$ are its mean and standard deviation over all 
	trials of the study.
	\autoref{tab:results} reports each model's mean and standard deviation of the effort index, which by construction has mean zero over all 2{,}100 trials.
}%
\ok{%
	We report Spearman $\rho$ between each process marker and rated creativity, pooled over all drawings and as the mean of the per-model correlations, in \autoref{tab:processproduct} of Appendix~\ref{app:processproduct}. 
}%


\subsection{Scorer Validation}%
\label{sec:scorer-validation}%
%
\textbf{AuDrA.}
\ok{%
	We first consider AuDrA by \citet{patterson2024audra} for scoring the creativity of our agent-generated drawings.
	AuDrA reaches Pearson $r = .80$ against human ratings on held-out human drawings.
	%
	However, agent drawings are potentially a far generalization of this corpus, and AuDrA already loses accuracy under a task shift within human drawings.
	On its far-generalization set (drawings from the object-transformation task), AuDrA falls to $r = .49$~\citep{patterson2024audra}.
	Following \citet{patterson2024audra}, we compute each drawing's inked-pixel count as an ink baseline. 
	On agent drawings, AuDrA's score correlates with this inked-pixel baseline 
	at Spearman $\rho = \knAudraInkRhoAgent$, against $\rho = \knAudraInkRhoHuman$ on human drawings.
	At this level of correlation, AuDrA scores agent drawings largely by elaboration, and we do not adopt it as our creativity measure.
}%

\IfFileExists{figures/scorer_validation.pdf}{%
	\begin{figure}[!htb]
		\centering
		\includegraphics[width=\linewidth]{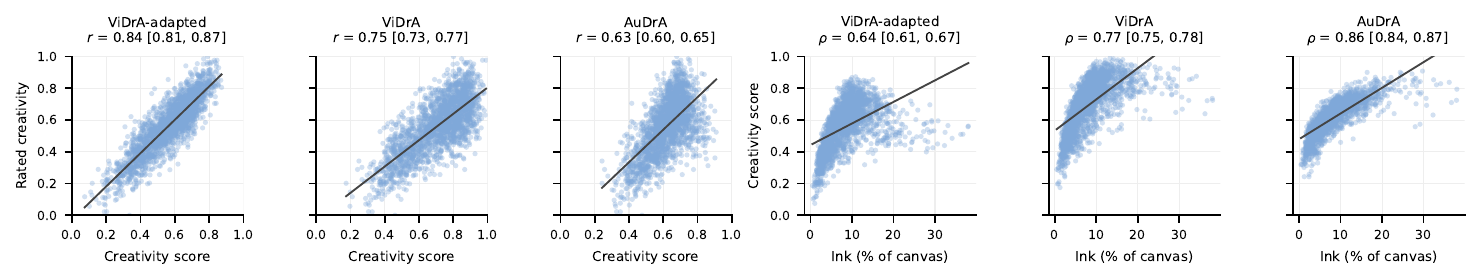}%
		\caption{%
			\ok{%
				Scorer validation over the agent drawings. Left three panels: each scorer's creativity score against rated creativity (Pearson's $r$).
				The ViDrA-adapted correlation is computed over its held-out test split.
				Right three panels: each scorer's creativity score against the inked-pixel baseline (Spearman's $\rho$).
				Lines are least-squares fits.}%
		}%
		\label{fig:scorer-validation}%
	\end{figure}%
}{}

\textbf{ViDrA.}
\ok{%
	We train an automated scorer, \textbf{ViDrA}, a kernel ridge regression (RBF kernel, $\alpha = 0.1$, $\gamma = 10^{-5}$) on frozen DINOv2 ViT-L/14 features~\citep{oquab2024dinov2},
	fit on the 11{,}075 rated human drawings from the public AuDrA corpus \citep{patterson2024audra}.
	ViDrA predicts creativity scores on the same normalized scale as AuDrA.
	%
	Each drawing is 
	resized to $448 \times 448$ and normalized with the ImageNet channel statistics.
	%
	We randomly partition the primary subset of the AuDrA corpus into $70/10/20$ train, validation, and test portions, 
	select settings on the validation split, and report 
	final results as mean and standard deviation over the test split.
	ViDrA reaches $r = \knVidraTestR \pm \knVidraTestRSd$ on this test split of human drawings, against AuDrA's published $r = .80$.%
}%

\textbf{ViDrA-adapted.}
\ok{%
	ViDrA is 
	fit on AuDrA's corpus of human drawings, and, like AuDrA, still correlates with inked-pixel baseline on agent drawings ($\rho = \knVidraInkRhoAgent$).
	To address this correlation, we adapt ViDrA by refitting its regression head on rated creativity, under leave-one-model-out and leave-one-stimulus-out cross-validation.
	The adapted head, \textbf{ViDrA-adapted}, reaches $r = \knAdaptModelR$ under the first scheme and $r = \knAdaptTaskR$ under the second, and it correlates with the inked-pixel baseline on agent drawings at $\rho = \knAdaptModelInkRho$.
	\autoref{fig:scorer-validation} shows each scorer's score against rated creativity and the inked-pixel baseline, and Table~\ref{tab:scorervalidation} in Appendix~\ref{app:scorer-validation} reports the correlations with 95\% confidence intervals.
}%


\subsection{Model Performance}%
\label{sec:performance}



\begin{figure}[htb]%
		\centering%
		\resizebox{.9\linewidth}{!}{%
			\includegraphics[width=\linewidth]{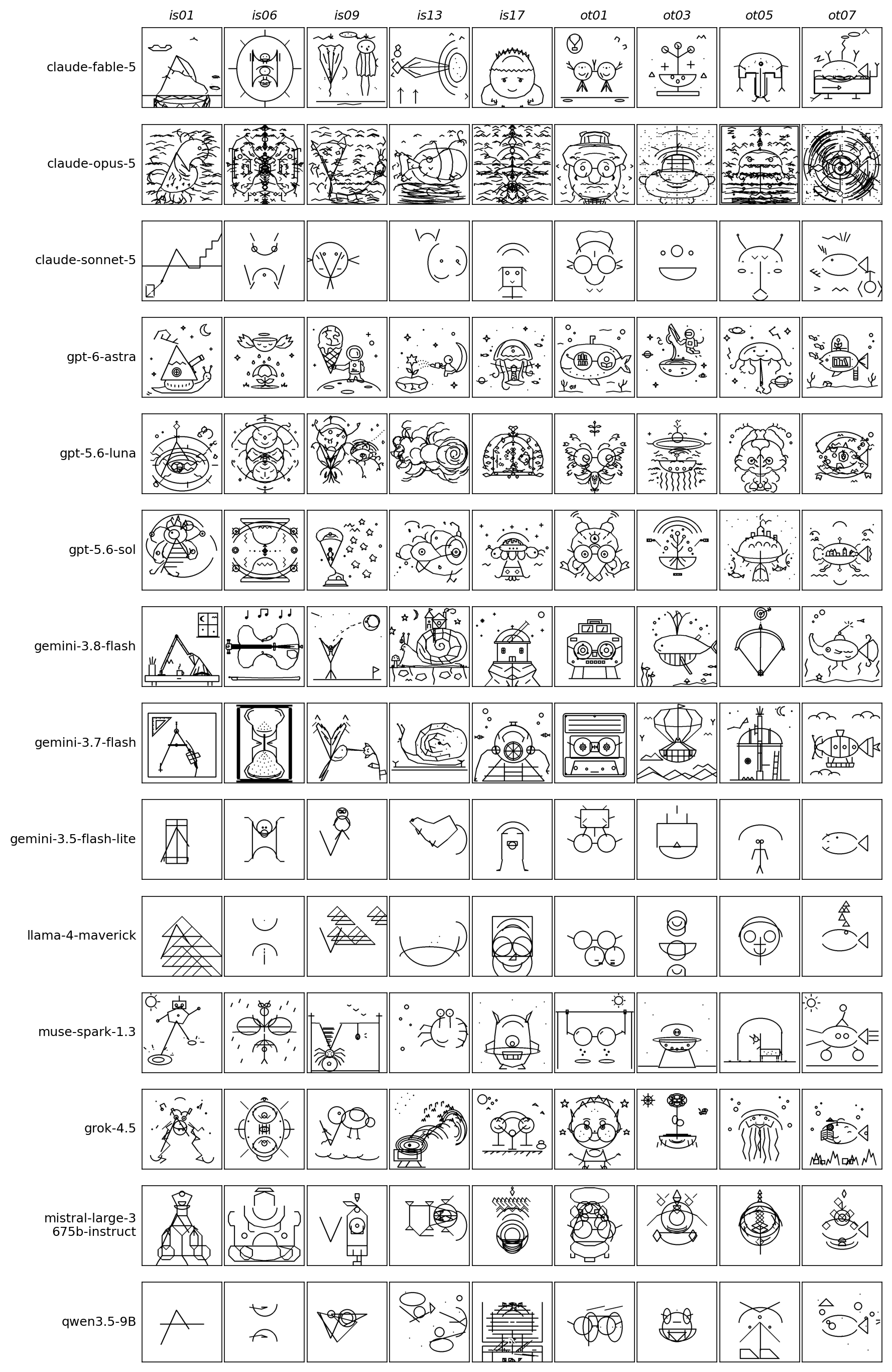}
		}%
		\caption{Example drawings from all evaluated models.
		}
		\label{fig:examples-full}
	\end{figure}

\ok{%
	\textbf{Agents score above the human mean on creativity but below it on recognizability.}
	Table~\ref{tab:results} summarizes the per-model results.
	Example drawings are displayed in \autoref{fig:examples} and \autoref{fig:examples-full}.
	Mean rated creativity is \knAgentMean{} over the agent drawings against \knHumanMean{} over the \knNHumanDrawings{} human reference drawings (difference \knAgentMinusHuman,
	95\% CI \knAgentMinusHumanCI, $d = \knAgentHumanD$).
	For eleven 
	of the 14 models, mean rated creativity exceeds the human mean.
	The mean agent drawing falls at the \knAgentPercentileOfHuman{} percentile of the human distribution. 
	On recognizability, however, the agent drawings score below the human reference drawings.
	Mean rated recognizability of agent drawings is \knAgentMeanRecog{} compared to \knHumanMeanRecog{} for the human reference drawings (difference \knAgentMinusHumanRecog, 95\% CI \knAgentMinusHumanRecogCI, $d = \knAgentHumanDRecog$).
	Mean rated recognizability exceeds the human mean in
	only four 
	of the 14 models, and
	the mean agent drawing falls at the \knAgentPercentileOfHumanRecog{} percentile of the human drawing distribution.
}%

\textbf{Agentic drawings fall far from the human distribution.}
\ok{%
	{%
		\looseness=-1
		Distance from human corpus is a drawing's mean cosine distance to its ten nearest corpus neighbors in the DINOv2 feature space, expressed as standard deviations above the median of the human drawings' own leave-one-out distances to the corpus.
		Every model's mean distance to the human corpus is at least 1.9~SD, and Gemini 3.8 Flash and Gemini 3.7 Flash sit at 4.3~SD (Table~\ref{tab:results}).
		AuDrA correlates with the inked-pixel baseline at Spearman $\rho = \knAudraInkRhoAgent$ on the agent drawings, against $\rho = \knAudraInkRhoHuman$ on the human drawing corpus it was trained on.
		This gap is expected when the agent drawings are a far generalization of AuDrA's human drawings. 
		Eleven of the 14 models place more ink on the canvas than the mean human drawing, from $+19.8$\% to $+268.3$\% (Claude Opus 5), and three models ink less.
		Claude Opus 5's ink coverage yields drawings qualitatively different from those of other models in the Claude family (\autoref{fig:examples}).%
	}%
}%

\newlength{\examplepanelsize}
\setlength{\examplepanelsize}{1.05cm}
\begin{figure}[!htb]
	\centering
	\setlength{\tabcolsep}{2pt}
	\small
\begin{tabular}{r@{\hskip 5pt}cccccccc}
	&
	\scriptsize\textit{is01} &
	\scriptsize\textit{is06} &
	\scriptsize\textit{is09} &
	\scriptsize\textit{is13} &
	\scriptsize\textit{is17} &
	\scriptsize\textit{ot01} &
	\scriptsize\textit{ot03} &
	\scriptsize\textit{ot05}
	\\[3pt]
	\scriptsize \mdlFable &
	\fbox{\includegraphics[width=\examplepanelsize]{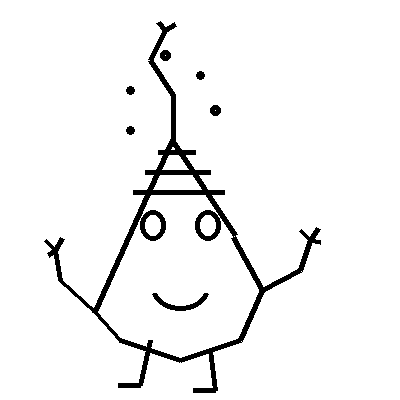}} &
	\fbox{\includegraphics[width=\examplepanelsize]{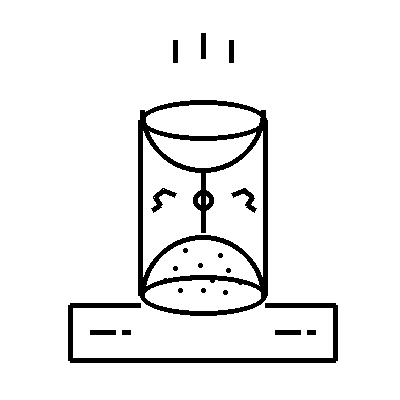}} &
	\fbox{\includegraphics[width=\examplepanelsize]{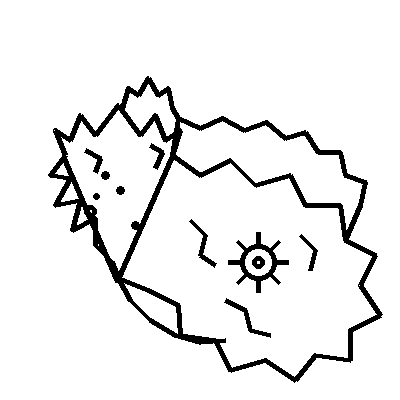}} &
	\fbox{\includegraphics[width=\examplepanelsize]{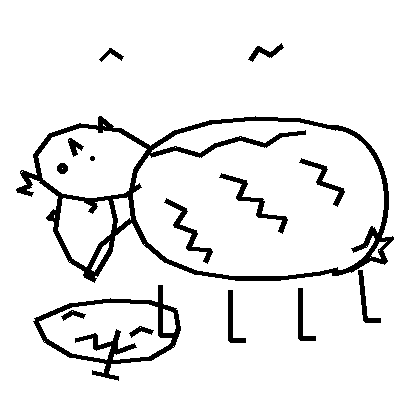}} &
	\fbox{\includegraphics[width=\examplepanelsize]{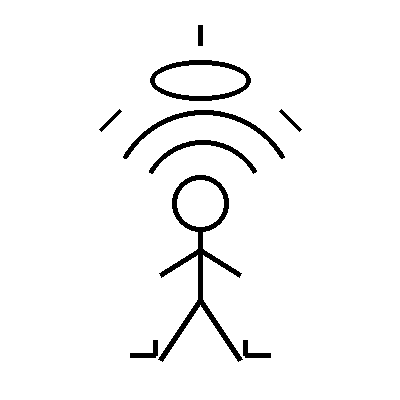}} &
	\fbox{\includegraphics[width=\examplepanelsize]{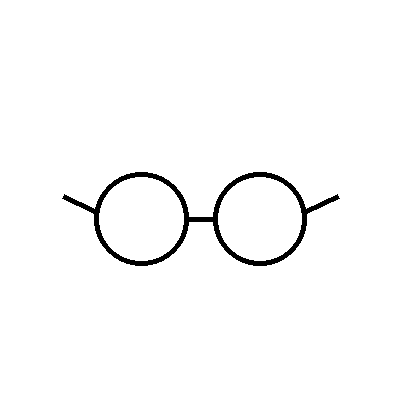}} &
	\fbox{\includegraphics[width=\examplepanelsize]{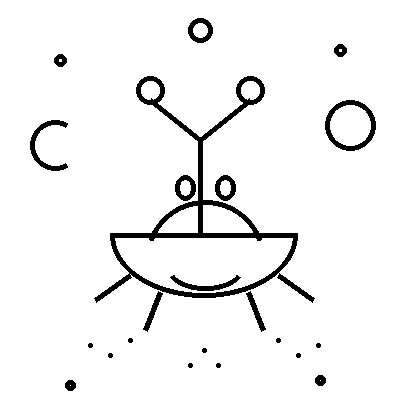}} &
	\fbox{\includegraphics[width=\examplepanelsize]{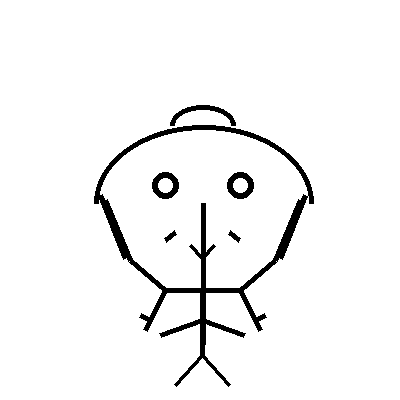}} \\[1pt]
	\scriptsize \mdlOpus &
	\fbox{\includegraphics[width=\examplepanelsize]{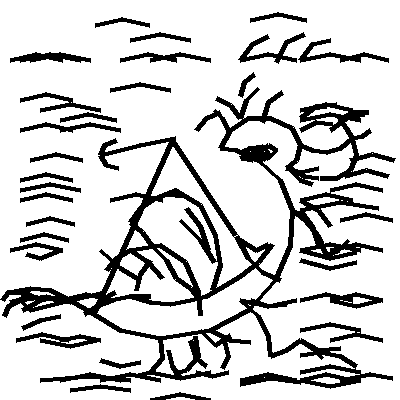}} &
	\fbox{\includegraphics[width=\examplepanelsize]{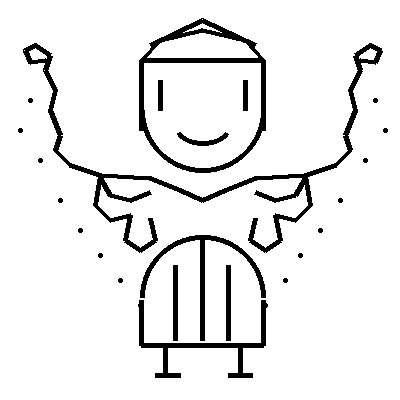}} &
	\fbox{\includegraphics[width=\examplepanelsize]{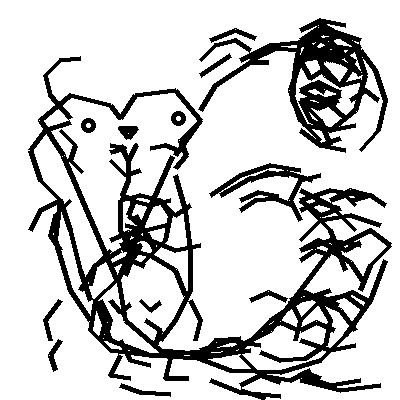}} &
	\fbox{\includegraphics[width=\examplepanelsize]{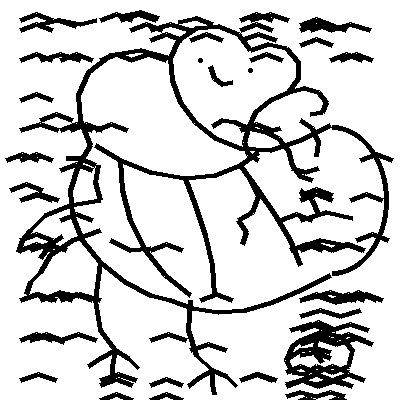}} &
	\fbox{\includegraphics[width=\examplepanelsize]{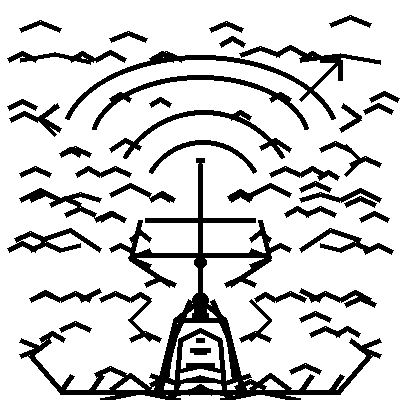}} &
	\fbox{\includegraphics[width=\examplepanelsize]{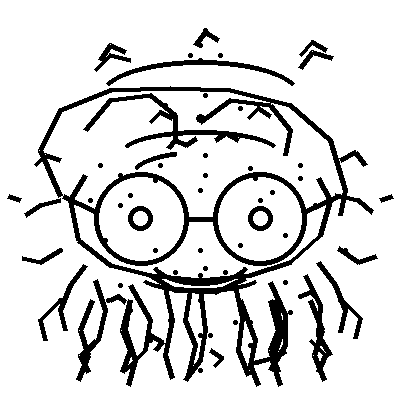}} &
	\fbox{\includegraphics[width=\examplepanelsize]{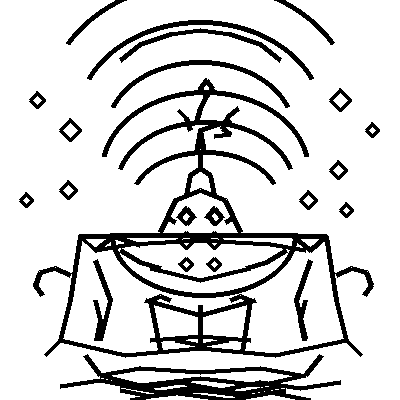}} &
	\fbox{\includegraphics[width=\examplepanelsize]{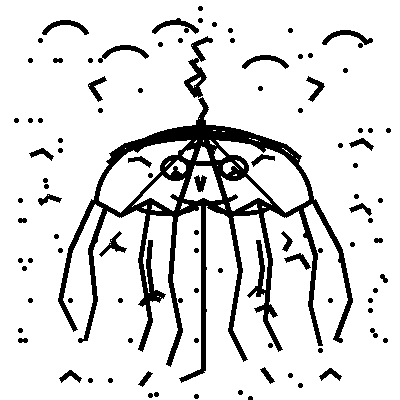}} \\[1pt]
	\scriptsize \mdlSonnet &
	\fbox{\includegraphics[width=\examplepanelsize]{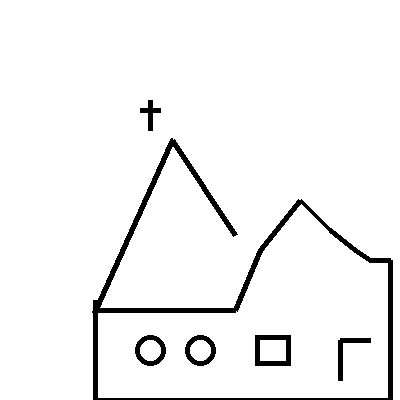}} &
	\fbox{\includegraphics[width=\examplepanelsize]{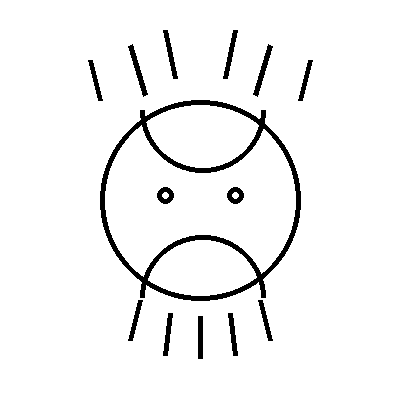}} &
	\fbox{\includegraphics[width=\examplepanelsize]{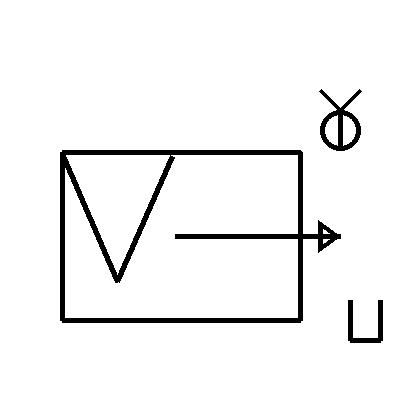}} &
	\fbox{\includegraphics[width=\examplepanelsize]{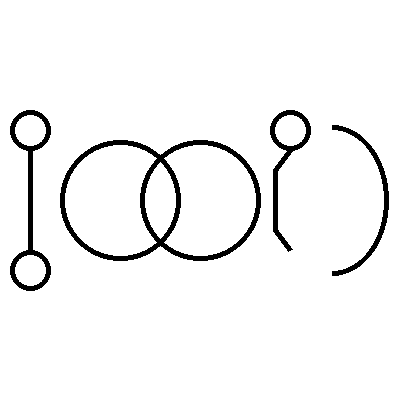}} &
	\fbox{\includegraphics[width=\examplepanelsize]{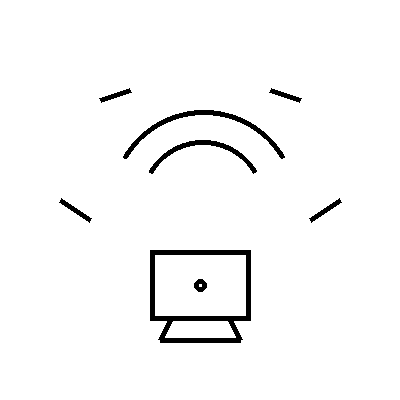}} &
	\fbox{\includegraphics[width=\examplepanelsize]{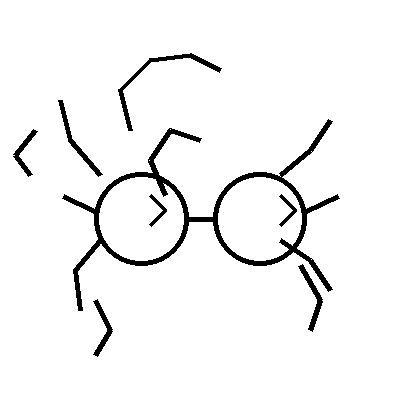}} &
	\fbox{\includegraphics[width=\examplepanelsize]{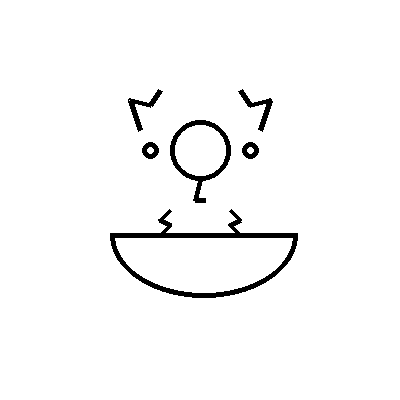}} &
	\fbox{\includegraphics[width=\examplepanelsize]{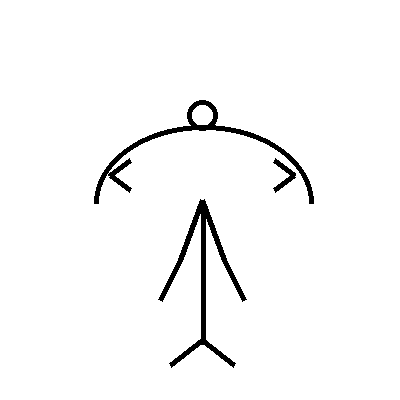}} \\[1pt]
\end{tabular}
	\caption{%
		\ok{%
			Example drawings 
			across 
			five incomplete shapes 
			and three object-transformation items 
			for three capability tiers of Anthropic's Claude.
		}%
	}%
	\label{fig:examples}
\end{figure}



\IfFileExists{tables/results_table_note.tex}{
\newcommand{\resultsnote}{ A percentile row states the share of human reference drawings that score below the model's mean composite. The human column reports the 300 human reference drawings, rated in the same sessions as the agent drawings. Its scorer cells are each scorer's mean over the human pool. Its remaining measures describe agent trials.}
}{}
\providecommand{\resultsnote}{}
\begin{table}[thb]
	\centering
	\caption{Primary evaluation results.
	}
	\label{tab:results}
	\small
	\renewcommand{\arraystretch}{0.95}%
\resizebox{\linewidth}{!}{%
\setlength{\tabcolsep}{3pt}%
\begin{tabular}{lccccccccccccccc}
\toprule
Metric
  & \rotatebox{90}{\shortstack[l]{human baseline\\(N=300)}}
  & \rotatebox{90}{Claude Fable 5}
  & \rotatebox{90}{Claude Opus 5}
  & \rotatebox{90}{Claude Sonnet 5}
  & \rotatebox{90}{GPT-6 Astra}
  & \rotatebox{90}{GPT-5.6 Luna}
  & \rotatebox{90}{GPT-5.6 Sol}
  & \rotatebox{90}{Gemini 3.8 Flash}
  & \rotatebox{90}{Gemini 3.7 Flash}
  & \rotatebox{90}{Gemini 3.5 Flash Lite}
  & \rotatebox{90}{Llama 4 Maverick}
  & \rotatebox{90}{Muse Spark 1.3}
  & \rotatebox{90}{Grok 4.5}
  & \rotatebox{90}{\shortstack[l]{Mistral Large 3 675B\\Instruct}}
  & \rotatebox{90}{Qwen3.5-9B}
\\
\midrule
Rated creativity $\uparrow$ & 0.41 & 0.52 & 0.56 & 0.42 & \textbf{0.81} & 0.61 & 0.68 & 0.71 & 0.67 & 0.35 & 0.40 & 0.59 & 0.60 & 0.55 & 0.37 \\
\quad $M$ on 1--5 scale $\uparrow$ & 2.39 & 2.79 & 2.95 & 2.43 & \textbf{3.95} & 3.12 & 3.45 & 3.55 & 3.40 & 2.12 & 2.33 & 3.11 & 3.12 & 2.94 & 2.21 \\
\quad $SD$ & 0.57 & 0.56 & 0.47 & 0.50 & 0.34 & 0.47 & 0.41 & 0.36 & 0.37 & 0.44 & 0.48 & 0.41 & 0.42 & 0.41 & 0.57 \\
\quad Percentile of human sample $\uparrow$ & -- & 77.3 & 83.3 & 57.3 & \textbf{99.7} & 89.7 & 96.7 & 97.3 & 96.0 & 35.7 & 52.0 & 89.0 & 89.3 & 83.0 & 39.3 \\
Rated recognizability $\uparrow$ & 0.53 & 0.45 & 0.37 & 0.39 & \textbf{0.73} & 0.33 & 0.44 & 0.70 & 0.65 & 0.33 & 0.32 & 0.55 & 0.50 & 0.30 & 0.25 \\
\quad $M$ on 1--5 scale $\uparrow$ & 3.21 & 2.85 & 2.41 & 2.50 & \textbf{4.14} & 2.21 & 2.75 & 3.99 & 3.75 & 2.24 & 2.15 & 3.32 & 3.04 & 2.06 & 1.88 \\
\quad $SD$ & 0.95 & 0.76 & 0.70 & 0.86 & 0.45 & 0.56 & 0.64 & 0.62 & 0.73 & 0.73 & 0.60 & 0.66 & 0.75 & 0.41 & 0.63 \\
\quad Percentile of human sample $\uparrow$ & -- & 37.3 & 24.3 & 26.7 & \textbf{80.3} & 20.3 & 34.3 & 75.3 & 64.7 & 20.3 & 18.7 & 48.0 & 41.7 & 15.3 & 11.3 \\
\midrule
ViDrA-adapted score $\uparrow$ & 0.39 & 0.53 & 0.57 & 0.43 & \textbf{0.78} & 0.62 & 0.68 & 0.70 & 0.67 & 0.36 & 0.39 & 0.58 & 0.60 & 0.56 & 0.39 \\
ViDrA score $\uparrow$ & 0.52 & 0.67 & 0.83 & 0.56 & \textbf{0.86} & 0.84 & 0.84 & 0.82 & 0.80 & 0.48 & 0.49 & 0.71 & 0.75 & 0.69 & 0.52 \\
AuDrA score $\uparrow$ & 0.49 & 0.59 & \textbf{0.77} & 0.52 & 0.69 & 0.72 & 0.68 & 0.68 & 0.67 & 0.48 & 0.51 & 0.59 & 0.63 & 0.64 & 0.52 \\
\midrule
Process effort index, $\bar{z}$ & -- & 0.01 & 1.25 & -0.20 & -0.06 & -0.22 & 0.02 & 0.10 & -0.07 & -0.49 & -0.43 & 0.10 & 0.38 & -0.20 & -0.20 \\
\quad $SD$ & -- & 0.32 & 1.28 & 0.27 & 0.18 & 0.22 & 0.18 & 0.26 & 0.22 & 0.20 & 0.22 & 0.20 & 1.00 & 0.18 & 0.62 \\
Mean tool calls & -- & 14.9 & 72.7 & 9.4 & 5.4 & 6.5 & 6.1 & 9.8 & 7.0 & 5.6 & 5.8 & 9.7 & 17.0 & 9.5 & 13.2 \\
Mean drawing operations & -- & 32.8 & 160.9 & 15.5 & 47.0 & 70.5 & 50.7 & 71.1 & 52.3 & 10.9 & 8.7 & 38.8 & 73.9 & 20.9 & 34.1 \\
Mean drawing operations per round & -- & 2.0 & 2.2 & 1.6 & 7.5 & 9.5 & 7.2 & 6.3 & 6.2 & 1.6 & 1.5 & 3.5 & 3.3 & 2.1 & 19.6 \\
Mean rounds to completion & -- & 17.3 & 72.9 & 11.5 & 6.4 & 7.5 & 7.1 & 12.2 & 8.9 & 7.4 & 5.9 & 12.0 & 18.7 & 10.5 & 4.5 \\
Mean tool diversity & -- & 1.9 & 0.7 & 1.8 & 2.2 & 1.5 & 2.3 & 2.1 & 2.0 & 1.4 & 1.6 & 2.4 & 2.4 & 1.8 & 1.8 \\
\midrule
Ink relative to human (\%) & -- & +23.1 & +268.3 & -18.7 & +97.7 & +130.0 & +85.1 & +137.0 & +122.6 & -44.9 & -2.1 & +19.8 & +57.7 & +82.8 & +23.2 \\
Distance from human corpus (SD) & -- & 2.1 & 3.9 & 2.3 & 3.5 & 3.3 & 3.0 & 4.3 & 4.3 & 1.9 & 3.0 & 2.9 & 2.7 & 3.4 & 2.8 \\
\midrule
Mean completion tokens & -- & 1734.9 & 7219.4 & 930.9 & 2333.6 & 2761.5 & 2253.7 & 22269.6 & 5629.7 & 407.1 & 519.6 & 22546.4 & 1955.3 & 1192 & 3365.3 \\
Mean time to completion (s) & -- & 83.6 & 271.3 & 24.8 & 47.4 & 46.8 & 48.7 & 125.5 & 54.3 & 13.7 & 24.6 & 363.2 & 208.8 & 150.7 & 30.1 \\
\midrule
Mean undo calls (\%) & -- & 6.32 & 0.00 & 7.20 & 0.06 & 0.00 & 0.00 & 8.94 & 7.18 & 6.99 & 0.00 & 9.67 & 3.86 & 0.00 & 0.12 \\
Mean erase calls (\%) & -- & 2.26 & 0.11 & 1.19 & 0.39 & 0.00 & 0.00 & 1.38 & 1.01 & 0.07 & 0.13 & 7.57 & 6.45 & 0.34 & 0.93 \\
\midrule
Finish rate (\%) $\uparrow$ & -- & \textbf{100.0} & \textbf{100.0} & \textbf{100.0} & \textbf{100.0} & \textbf{100.0} & \textbf{100.0} & \textbf{100.0} & \textbf{100.0} & \textbf{100.0} & \textbf{100.0} & \textbf{100.0} & \textbf{100.0} & \textbf{100.0} & 99.3 \\
Failed tool calls (\%) $\downarrow$ & -- & 0.04 & \textbf{0.00} & 0.06 & \textbf{0.00} & 0.18 & \textbf{0.00} & \textbf{0.00} & \textbf{0.00} & \textbf{0.00} & \textbf{0.00} & 0.06 & \textbf{0.00} & \textbf{0.00} & 0.39 \\
Malformed operations (\%) $\downarrow$ & -- & 0.11 & 0.01 & \textbf{0.00} & \textbf{0.00} & 0.35 & 0.09 & \textbf{0.00} & \textbf{0.00} & \textbf{0.00} & \textbf{0.00} & 0.39 & 0.57 & 0.83 & 23.91 \\
\bottomrule
\end{tabular}
}%

\end{table}

\textbf{Rated creativity separates the models.}
\ok{%
	{%
		\looseness=-1
		\modelname{\knBestModel} attained the highest rated creativity (mean \knBestModelMean, 95\% CI \knBestModelCI) and \modelname{\knWorstModel} the lowest (mean \knWorstModelMean, 95\% CI \knWorstModelCI). 
		In a linear mixed-effects regression that predicts each drawing's rated creativity from the model that drew it and the stimulus type (incomplete shape or object transformation), with a random intercept for each of the 30 stimuli, the model effect is significant (Wald $\chi^2(\knModelEffectDf) = \knModelEffectChiSq$, $p \knModelEffectP$) and the variance attributable to stimuli is near zero.
	}%
}%

\textbf{The adapted ViDrA reproduces the rated model ranking on held-out models.}
\ok{%
	\looseness=-1
	Under the leave-one-model-out scheme (Section~\ref{sec:scorer-validation}), each model 
	is scored by a head fit on the other 13 models' ratings.
	These held-out scores rank the models nearly identically to the creativity ratings (
	$\rho = \knAdaptModelRankRho$). 
}%

\textbf{Creativity and recognizability are correlated but distinct.}
\ok{%
	The two ratings correlate at $r = \knCreatRecogR$ (95\% CI \knCreatRecogCI) over the 
	agent drawings and at a mean of $\bar{r} = \knCreatRecogWithinR$ (95\% CI \knCreatRecogWithinCI) within models.
	Across the 
	agent drawings, rated creativity correlates with the inked-pixel count at $r = \knCreativityInkR$ and recognizability at $r = \knRecognizabilityInkR$.
	The \knHumanRatingInkN{} human 
	drawings show the same pattern, $r = \knHumanCreativityInkR$ for creativity and $r = \knHumanRecognizabilityInkR$ for recognizability.
}%

\ok{%
	\textbf{Models differ widely in drawing effort.}
	The median trial took 8 rounds and 9 tool calls.
	Half of all trials finish within 6 to 13 rounds, but the distribution of rounds has a long tail (mean 14.5, SD 21.3).
	The longest trial, from Claude Opus 5, ran 261 rounds with a total of 992 drawing operations.
	The median drawing is assembled from 37 drawing operations at about 3 operations per round, while the extreme is a 2{,}178-operation drawing by Grok 4.5.
	All trials were ended by the agent, with the exception of one Qwen3.5-9B trial in which the model returned no tool call.
}%

\ok{%
	\textbf{Models concentrate their tool use on lines and polylines.}
	Lines and polylines account for 67\% of all drawing operations.
	Lines are the most used tool type in eight of the 14 models, and polylines in five models.
	Gemini 3.5 Flash Lite draws 80\% of its operations as lines, Claude Opus 5 draws 68\% as polylines, and Grok 4.5 is the only model whose most used tool type is dots.
	Chords, pie slices, regular polygons, and rounded rectangles together account for under 1\% of drawing operations, and each of these four tools is used by 4 to 10 of the 14 models.
	Mean tool diversity (the Shannon entropy of a trial's tool-type distribution) ranges from 0.7 to 2.4 across models (Table~\ref{tab:results}).
}%

\ok{%
	{%
		\looseness=-1
		\textbf{Revision is rare.}
		Undo appears in \knUndoTrials{} trials (\knUndoTrialsPct) and erase in \knEraseTrials{} trials (\knEraseTrialsPct).
		Undo splits the models into two groups (\autoref{tab:results}).
		Seven 
		models each invoked undo in at least \knUndoActiveMin{} of their 150 trials.
		\knUndoTopModel{} used the revision tools most, with undo in \knUndoTopModelUndo{} trials (70\%) and erase in \knUndoTopModelErase{} trials (50\%), while 
		seven models invoked undo in at most \knUndoOtherMax{} trials and 
		five never used undo.
		The median revising trial has 2 revision calls, and trials that spend at least 30\% of their calls on revision are rare (77 trials; 3.7\%).
		Failed tool calls occur in \knToolErrorTrials{}~trials (0.5\%) and malformed operations in \knMalformedTrials{}~trials (4.1\%).
		\knErrorTopModel{} accounts for \knErrorTopModelToolErrors{} of the trials with failed calls and \knErrorTopModelMalformed{} with malformed operations.
	}%
}%

\ok{%
	\textbf{Drawing effort separates models, not trials.}
	Pooled over all 2{,}100 trials,
	drawing operations per round 
	correlates with rated creativity at $\rho = \knPPOpsPerRoundPooledRho$ (95\% CI \knPPOpsPerRoundPooledCI).
	Within models, the mean correlation falls to $\bar{\rho} = \knPPOpsWithinRho$ (95\% CI \knPPOpsWithinCI). 
	The number of tool calls is close to uncorrelated with rated creativity ($\rho = \knPPCallsPooledRho$ pooled, $\bar{\rho} = \knPPCallsWithinRho$ within models).
	Correlations of the remaining process markers with rated creativity are reported in Table~\ref{tab:processproduct} of Appendix~\ref{app:processproduct}.%
}%
%

\label{app:examples}

%
%
%
%
%
%
%
	%
%
%
\subsection{Failure analysis}%
\label{sec:failure-analysis}%
%
\ok{%
	We reviewed the final drawings 
	on per-model contact sheets (grids of all final drawings),
	and analyzed 
	patterns and failures qualitatively, without predefined categories. 
	For trials that stood out, we compared the final canvas with the per-call snapshots and reviewed the tool-call and reasoning traces.
	The failures aggregate into five 
	modes.%
}%

\ok{%
	\textit{Absent or unproductive revision.}
	After every round, the agent has an opportunity to inspect the canvas and issue erase and undo calls.
	Yet, most trials neither erase nor undo (Section~\ref{sec:performance}).
	When a model does revise, the revision sometimes fails to converge.
	Some trials (e.g., in Claude Fable~5 and Sonnet~5) alternate drawing and undoing.
	One constraint of the harness is that undo reverts the entire call.
	Models batch multiple operations into one call, and
	one misplaced stroke, thus, may cost the whole batch if it is undone.
	Gemini 3.8 Flash works within this constraint by undoing the full batch and redrawing an adjusted copy. This loop is costly, but does converge.
	Grok 4.5 revises by erasing, but it can fail to stop.
	In one of Grok's trials, 
	most rounds are erase calls that remove some stray marks and leave others, and the final canvas still carries stray marks.
}

\ok{%
	\textit{Null responses.}
	Eleven trials end with fewer than 300 added inked pixels, nine with none. Five of these trials come from Claude Fable~5 and four from Qwen3.5-9B.
	In one trial (Claude Fable 5, see stimulus ot01 in \autoref{fig:examples}),
	the model starts drawing, but then undoes all rounds and declares the drawing finished
	with no added ink over the starting stimulus.
	Null responses arise in two ways.
	In the first, the model gives up after one or several draw-then-retract loops, as in the trial above.
	In the second, every operation the model submits is malformed and skipped, as in one Qwen3.5-9B trial titled ``Fish Shimmering with Geometric Movement'' whose canvas never changes.
}%

\ok{%
	\textit{Overinking and effort without creativity gain.}
	Claude Opus 5 draws 
	more than any other model, with a median of 66.5 rounds 
	and a median of 27{,}800 added inked pixels (17.4\% of the canvas). 
	The model typically places a central figure within the first quarter of its trial's rounds, but the remaining rounds are spent filling the background with repeated texture strokes and hatching. 
	For instance, in one Opus 5 trial (stimulus is20, replicate 5), a nesting bird is completed early and about 120 further rounds add rain texture until it crowds the figure. 
	Claude Opus 5's mean rated creativity ranks eighth of the 14 models, and within its trials added ink correlates negatively with rated creativity ($r = -.21$) and recognizability ($r = -0.48$).
	GPT-5.6 Luna overinks with ornaments rather than textures, often wrapping central figures in symmetric flourishes.
	The model pairs one of the largest median ink budgets with a mean recognizability of only 0.33 (fifth-lowest of the 14 models).
}%

\ok{%
	\textit{Homogenization 
		across independent trials.}
	Trials run independently with no shared context, yet each model returns to a small set of ideas and themes (cf. Appendix \ref{app:titles}).
	GPT-5.6 Luna titles 92 (61.3\%) of its 150 drawings ``cosmic''. 
	Mistral Large 3 draws a robot in 75 trials (50\%), Claude Sonnet 5 a face in 58 trials (38.7\%), and GPT-6 Astra a snail in 36 trials (24\%). 
	Each drawing is rated individually, so 
	rated creativity does not register this collapse of the response distribution.
}%

\ok{%
	\textit{Fragmentary drawings.}
	The lowest-rated models add a few disconnected primitives that neither form a coherent object nor connect to the stimulus in meaningful ways.
	Qwen3.5-9B and Llama 4 Maverick scatter circles, triangles, and zigzag lines.
	Qwen3.5-9B also loses ink to 
	malformed tool calls, with 606 drawing operations rejected for schema errors, such as repeated coordinate names.
	While the harness reports each rejection in its tool response in the following round and the canvas image shows no change, the model continues 
	without correcting the error.%
}%
%
%
%
%
%
%
%
%
\vspace{-.9\baselineskip}
\subsection{Sensitivity Analyses}%
\label{sec:sensitivity-analyses}%
\label{sec:ablations}%
\vspace{-.5\baselineskip}
%
\ok{%
	We examine how design choices in the benchmark harness affect 
	agent performance.
	Each analysis is an ablation-style manipulation of one harness design choice.
	The six manipulations (see Table~\ref{tab:sweeps}) run the full 30-stimulus bank with one anchor model (GPT-5.6 Luna). 
	As in the main study, we collect 12 crowdsourced ratings of creativity and recognizability per drawing.
	For each manipulation, we report the mean difference between the manipulated and its baseline condition (Table~\ref{tab:sweeps}), paired by stimulus, with a 95\% bootstrap confidence interval.
}%

\ok{%
	The \textit{interface manipulation} replaces the drawing tools with one tool that takes SVG markup, a representation likely familiar from pretraining.
	Mean rated creativity is \ablSvgCondMean{} under the SVG interface and \ablSvgBaseMean{} 
	under tool calling,
	a paired difference of \ablSvgDelta{} (95\% CI \ablSvgCI, $d$~$=$ \ablSvgDeltaSd{}).
	Appendix~\ref{app:svg-ablation} shows examples of the SVG drawings and reports a more detailed comparison.
}%


\ok{%
	The \textit{vocabulary manipulation} compares the {full} set of tools against a {minimal} set, which consists of only \texttt{draw\_dots}.
	Restricting the vocabulary lowers mean rated creativity from \ablPrimitivesBaseMean{}
	to \ablPrimitivesCondMean{}, a paired difference of \ablPrimitivesDelta{} (95\% CI \ablPrimitivesCI, $d$~$=$ \ablPrimitivesDeltaSd{}). 
}%


\ok{%
	{%
		\looseness=-1
		The \textit{blank-canvas manipulation} removes the starting stimulus, which separates
		difficulty with the incomplete-drawing task from difficulty with drawing itself.
		Mean rated creativity is \ablBlankCondMean{} without the stimulus against \ablBlankBaseMean{}
		at baseline, a difference of \ablBlankDelta{} (95\% CI \ablBlankCI, $d$~$=$ \ablBlankDeltaSd{}).
	}%
}%



\ok{%
	The \textit{oracle manipulation} names the object each stimulus is to become, drawn from a fixed list. This removes the choice of what to depict and leaves only the composition and the execution to the model.
	Naming the target raises mean recognizability from \ablOracleBaseMean{} in the anchor model's primary-study cells to \ablOracleCondMean{}, a paired difference of \ablOracleDelta{} (95\% CI \ablOracleCI, $d$~$=$ \ablOracleDeltaSd{}). 
}%


\vspace{-.8\baselineskip}%
\begin{table}[!tbh]%
	\centering%
	\caption{Sensitivity manipulations.}%
	\label{tab:sweeps}%
	\small%
	\setlength{\tabcolsep}{3pt}%
	\resizebox{\linewidth}{!}{%
		\begin{tabular}{@{}lllrrl}
			\toprule%
			Manipulation & Manipulated condition & Baseline & Repl. & Trials & Question
			\\
			\midrule
			Interface   & SVG markup & tool calls (primary study, 150) & 5 & 150 & Does the drawing interface affect rated creativity? \\
			
			Vocabulary  & minimal vocabulary & full vocabulary & 1 & 60 & Does the number of drawing tools affect rated creativity? \\
			
			Blank canvas& no stimulus & stimulus (framing baseline, 30) & 1 & 30 &
			Does the starting stimulus affect rated creativity? \\
			
			Oracle      & subject given & subject withheld (framing baseline, 30) & 1 & 30 & Does the choice of subject affect recognizability? \\
			
			Framing     & depictive, example & canonical (primary study's context) & 1 & 90 & Does the wording of the instruction affect recognizability? \\
			
			Context     & $2^3$ context components & 
			primary study's context
			& 1 & 240 & How does the per-turn context affect the rated creativity?
			\\
			\bottomrule%
		\end{tabular}%
	}%
\end{table}%

\ok{%
	{%
		\looseness=-1
		The \textit{framing manipulation} varies the task instruction at {canonical} (the wording of Appendix~\ref{app:prompts}), {depictive} (adding that the drawing must depict something a viewer could identify without being told what it is), and {example} (adding to that a worked example of transforming a shape into an object).
		Mean recognizability is \ablFramingDepictiveBaseMean{} under the canonical instruction, \ablFramingDepictiveCondMean{} under the depictive framing, and \ablFramingExampleCondMean{} under the example framing.
		Against the canonical instruction, the depictive framing shifts recognizability by \ablFramingDepictiveDelta{} (95\% CI \ablFramingDepictiveCI, $d$~$=$ \ablFramingDepictiveDeltaSd{}) and the example framing by \ablFramingExampleDelta{} (95\% CI \ablFramingExampleCI, $d$~$=$ \ablFramingExampleDeltaSd{}). 
	}%
}%


\ok{%
	The \textit{context manipulation} crosses the three components of the context (the tool-call history, current canvas, and strip of recent snapshots), each present or absent.
	The outcome for this manipulation is the ViDrA score over the 240 drawings of the sweep.
	The image of the current canvas raises the mean score from 0.77 to 0.84 (95\% CIs $[0.76, 0.79]$ and $[0.82, 0.86]$).
	The tool-call history does not move the mean, at 0.82 in both levels.
	The snapshot strip lowers the mean from 0.84 to 0.80 (95\% CIs $[0.82, 0.86]$ and $[0.79, 0.82]$).
	%
	The primary study's context \{history, canvas, strip\} trails the best condition, history, by 0.03 on creativity and 0.06 on recognizability, on their shared $[0,1]$ scale.%
}%

\ok{%
	The creativity level of Section~\ref{sec:performance} does not depend on the drawing interface, the vocabulary, the stimulus, or the per-turn context.
	Naming the subject predictably raises recognizability,
	showing that the model is capable of executing a recognizable drawing.
	PainterBench's open instruction separates drawing capability from creative judgment.
	The model can make its drawings recognizable but does not adopt recognizability as a goal unprompted, as human participants do.%
}%


\vspace{-.3\baselineskip}%
\section{Conclusion}
\label{sec:discussion}%
\vspace{-.5\baselineskip}%
%
We introduced PainterBench, which ports the 
figural divergent-thinking task to tool-using agents, and evaluated 14 multimodal language models on this task.
An automated creativity scorer does not transfer to agent drawings, so we compare models on ViDrA-adapted, our scorer refit on \rev{the training split of} crowdsourced ratings of agent drawings.
\knBestModel{} produces the most creative drawings, and rated creativity separates the models. 
The differences between models persist after accounting for the stimuli.

Under the standard definition of creativity, a creative product is original and effective~\citep{runco2012standard}.
The agent drawings exceed the human reference drawings on 
creativity, which we interpret as an advantage in originality, but fall below them on recognizability, which we interpret as a lack in effectiveness.
%
For the anchor model, the sensitivity analyses place the recognizability deficit in the goal the model pursues rather than in execution, since 
naming the subject brings recognizability within 0.03 of the human reference mean.
Revision of the canvas is rare, and added ink earns no gain in creativity.
Each model gravitates toward a small set of ideas across independent trials.

\textit{Limitations and future work.}
\label{sec:limitations}
\label{sec:future-work}
\ok{%
PainterBench measures originality, the component of creativity that figural divergent-thinking tests assess.
We used an approach that maintains compatibility with the MTCI task and the AuDrA corpus.
Whether the same agents draw more creatively with freehand strokes or color is untested. 
In addition, the sensitivity analyses run on one anchor model and bound the results for that model.
}%
%
%
\ok{%
Agent drawings lie far from the corpus of human drawings (Section~\ref{sec:scorer-validation}).
ViDrA initially learns from the human drawings, and ViDrA-adapted refits only the scoring head on the crowdsourced agent ratings.
Future work includes scoring further components of creativity, such as elaboration, flexibility, fluency, and aesthetic appeal.
A scorer trained natively on rated agent drawings is a second direction, and the released drawings, ratings, and benchmark harness supply the corpus to build one.
}%



\clearpage

\section*{AI use statement}%
%
In this work, we used generative AI tools for code development and for drafting early versions of some sections.
We have not used generative AI tools for research ideation, the study design, or the analyses, which are our own.
We have reviewed all AI-assisted work.
The authors verified all generated code, revised all generated text, and reviewed and revised the final manuscript in full.
We take responsibility for the final content of this work, including text, claims, and artifacts produced with the aid of generative AI.

\section*{Ethics statement}%
\ok{%
This study involved human participants in a low-risk annotation role:
\rev{1{,}254} unique crowdworkers on the CloudResearch platform rated machine-generated drawings on two 5-point questions, in batches of 30 drawings per worker.
The median completion time was 3 min 21 s,
and participants were compensated at \$13.5 per hour. 
Payment was calibrated in a small-scale pilot~\citep{oppenlaenderpilot}.
No personal information beyond the platform-assigned 
data was collected.
Participants were not informed that some images were created by AI. This was done for two reasons: to match AuDrA's task instruction \citep{patterson2024audra}, and to not bias the ratings \citep{10.1037/aca0000136,10.1145/3334480.3382892}.
The rated images are line drawings and contain no depictions of known people or sensitive content.
The low-risk study protocol is exempt from ethics review under our institutional and national guidelines.%
}%
%
%
\section*{Reproducibility statement}%
%
\ok{%
We release
all generated drawings (2{,}100 final drawings and 600~drawings from the sensitivity analyses)
as well as all \ok{35{,}004} per-round canvas snapshots,
the full logs of every run (chat history, tool-call trace from which all process markers are computed, and drawing titles),
the benchmark harness (including YAML configurations that specify every run),
the set of tools and system-prompt definitions for every sensitivity-analysis condition,
the 30 stimuli as deterministic procedural vector programs,
and the crowdsourced data with 72,000 ratings.
The system prompt is reported in Appendix~\ref{app:prompts} and the tool surface is specified in Appendix~\ref{app:tools}.
The main study and sensitivity analyses can be rerun with a single command per configuration file, including on future model versions.
However, provider sampling is not deterministic, so a rerun yields new drawings.
The annotation protocol, rating instrument, and agreement statistics are specified in Section~\ref{sec:eval} and included in the release, together with the analysis scripts.
ViDrA's training pipeline, its splits over the primary subset of the public AuDrA corpus, and the fitted checkpoint are also released.
For peer review, the agent drawings are supplied in the Supplementary Material.
The full release, including per-round canvas snapshots and ViDrA checkpoints, is available at \url{https://huggingface.co/datasets/painterbench/painterbench}.%
}%

%

\bibliographystyle{iclr2027_conference}
\bibliography{paper}

\newpage
\appendix

%
%
%
%

\section{Evaluated Models}
\label{app:models}

\noindent
\ok{%
Table~\ref{tab:models} lists the evaluated models with the exact API request identifier
and each model's release date.
}%

\begin{table}[htb]
\caption{Evaluated models.}
\label{tab:models}%
\centering%
\small
\IfFileExists{tables/models_table.tex}{
{%
\setlength{\tabcolsep}{3pt}%
\begin{tabular}{llll}
\toprule
Model & Provider & Identifier & Released \\
\midrule
GPT-6 Astra & OpenAI & openai/gpt-6-astra & 2026-09-04 \\
GPT-5.6 Sol & OpenAI & openai/gpt-5.6-sol & 2026-07-09 \\
GPT-5.6 Luna & OpenAI & openai/gpt-5.6-luna & 2026-07-09 \\
Claude Fable 5 & Anthropic & anthropic/claude-fable-5 & 2026-06-09 \\
Claude Opus 5 & Anthropic & anthropic/claude-opus-5 & 2026-07-24 \\
Claude Sonnet 5 & Anthropic & anthropic/claude-sonnet-5 & 2026-06-29 \\
Muse Spark 1.3 & Meta & meta/muse-spark-1.3 & 2026-09-02 \\
Mistral Large 3 675B Instruct & Mistral AI & mistralai/mistral-large-2512 & 2025-12-01 \\
Qwen3.5-9B & Alibaba & qwen/qwen3.5-9b & 2026-03-10 \\
Llama 4 Maverick & Meta & meta-llama/llama-4-maverick & 2025-04-05 \\
Gemini 3.8 Flash & Google & google/gemini-3.8-flash & 2026-09-02 \\
Gemini 3.7 Flash & Google & google/gemini-3.7-flash & 2026-08-13 \\
Gemini 3.5 Flash Lite & Google & google/gemini-3.5-flash-lite & 2026-07-21 \\
Grok 4.5 & xAI & xai/grok-4.5 & 2026-06-29 \\
\bottomrule
\end{tabular}
}%
}{}%
\end{table}%
\FloatBarrier

\section{PainterBench Instruction and System Prompt}
\label{app:prompts}

\ok{\textit{Initial user instruction:}}



\begin{quote}
\footnotesize\ttfamily
The canvas shows the starting shape. Create the most original drawing you can think of.
\end{quote}


\ok{\textit{System prompt}:} 

\begin{quote}\footnotesize\ttfamily
You are taking a figural creativity test on a digital canvas of size 400x400.\\
The canvas already contains a black starting shape.\\[4pt]
Task:\\
Create the most original drawing you can think of.\\
The starting shape must be incorporated as part of your drawing. It is part of the canvas and cannot be erased. Draw the most original drawing you can think of that uses it.\\[4pt]
Rules:\\
Communicate ONLY by calling tools.\\
Issue exactly one tool call per response.\\
Each drawing tool accepts a list of operations: batch all shapes of the same type into a single call, then use a separate call for a different shape type.\\
A snapshot is saved automatically after every drawing call.\\
When the drawing is finished, call drawing\_finished(label=...) with a short title describing what you drew.\\
Never ask questions. Do not request confirmation. Assume you should continue unless you are done.\\[4pt]
All strokes are black on a white canvas. To erase, call any drawing tool with erase=true.\\
It lays white along exactly the path that shape would have drawn, so an erase costs the\\
same stroke the drawing did. The starting shape reappears if you erase over it.\\
Important: every stroke uses a fixed width of 5, drawing and erasing alike.\\
Do not try to vary line thickness.\\
Every shape is drawn as an outline. To ink a solid area, draw the strokes that cover it.\\
Note that you cannot layer; when you draw something, you will draw on top of existing drawn things.\\
If the last tool call did not land as intended, call undo\_last\_action() to revert it before issuing a new stroke.\\
Note: you will see your 10 most recent rounds of tool calls. Earlier rounds are omitted, so treat the canvas images as the record of what has been drawn.
\end{quote}

\ok{%
The sensitivity analyses of Section~\ref{sec:ablations} amend this prompt per condition.
The \emph{framing} manipulation's {depictive} condition inserts \texttt{Your drawing must depict something recognizable. Someone who sees the finished canvas without being told its title should be able to say what it shows.} at the end of the task section.
The \emph{framing} manipulation's {example} condition appends to that \texttt{For example, someone given a circle might draw a clock face, adding hands and numerals inside it, rather than drawing further circles beside it.}
The \emph{oracle} manipulation inserts \texttt{Draw <target>. Incorporate the starting shape into it. This replaces the instruction to choose your own subject; the subject is given.} at the same position, and the opening user message reads \texttt{Draw <target>.} instead of the canonical instruction.
The \emph{vocabulary} manipulation's {minimal} condition appends a line naming the only available drawing tool.
The \emph{interface} manipulation's {SVG} condition replaces the batch-tool rule with an instruction to draw by calling \texttt{draw\_svg} and appends a paragraph listing the supported SVG elements.
The \emph{context} manipulation's conditions replace the closing note with one that states which history the condition provides.
The \emph{blank-canvas} manipulation removes every sentence that mentions the starting shape.
}%

\section{Tool Definitions}%
\label{app:tools}%
%
\ok{%
This appendix specifies the tool surface summarized in Table~\ref{tab:tools}.
Appendix~\ref{app:signatures} gives the call signature of every tool, and Appendix~\ref{app:schema} the schema format in which a tool is presented to the model.
}%

\subsection{Call Signatures}
\label{app:signatures}
\ok{%
Every drawing tool takes a list of operations as its first argument 
and an optional \texttt{erase} flag as its second argument.
Table~\ref{tab:signatures} lists the arguments for each tool.
Fields marked with a question mark are optional.
Coordinates are integer pixel positions on the $400 \times 400$ canvas, with the origin at the top left.
Angles are in degrees, measured clockwise from three o'clock.
Every operation is in black and every operation is drawn with a fixed 5-pixel stroke to match the AuDrA dataset.
An operation whose fields cannot be parsed is skipped, and the count of skipped operations is returned to the agent. The remaining operations in the same call are still drawn.
}%

\begin{table}[htb]
\caption{Call signatures of the tools in the primary study.
A question mark indicates an optional argument or field.
}%
\label{tab:signatures}%
\centering%
\footnotesize
{%
\setlength{\tabcolsep}{3pt}%
\begin{tabular}{lll}
\toprule
Tool & Arguments & Fields of one operation \\
\midrule
draw\_dots & dots, erase? & x, y \\
draw\_lines & lines, erase? & x1, y1, x2, y2 \\
draw\_polylines & polylines, erase? & points, closed? \\
draw\_polygons & polygons, erase? & points \\
draw\_regular\_polygons & polygons, erase? & x, y, radius, n\_sides, rotation? \\
draw\_rectangles & rects, erase? & x1, y1, x2, y2 \\
draw\_rounded\_rectangles & rects, erase? & x1, y1, x2, y2, radius \\
draw\_circles & circles, erase? & x, y, radius \\
draw\_ellipses & ellipses, erase? & x1, y1, x2, y2 \\
draw\_arcs & arcs, erase? & x1, y1, x2, y2, start\_angle, end\_angle \\
draw\_pieslices & pieslices, erase? & x1, y1, x2, y2, start\_angle, end\_angle \\
draw\_chords & chords, erase? & x1, y1, x2, y2, start\_angle, end\_angle \\
\midrule
undo\_last\_action & --- & --- \\
\midrule
drawing\_finished & label & --- \\
\bottomrule
\end{tabular}
}%
%
\end{table}%
\FloatBarrier


\subsection{Schema Format}
\label{app:schema}

\ok{%
The harness presents each tool to the model as an OpenAI-style function schema.
For instance, the schema for \texttt{draw\_polygons} is:
}%

\ok{%
\begin{quote}\footnotesize\ttfamily
\{"type": "function", "function": \{\\
\hspace*{1em}"name": "draw\_polygons",\\
\hspace*{1em}"description": "",\\
\hspace*{1em}"parameters": \{"type": "object",\\
\hspace*{2em}"properties": \{\\
\hspace*{3em}"polygons": \{\\
\hspace*{4em}"type": "array",\\
\hspace*{4em}"items": \{"type": "object"\},\\
\hspace*{4em}"description": "List of \{points:[[x,y],...]\}"\},\\
\hspace*{3em}"erase": \{\\
\hspace*{4em}"type": "boolean",\\
\hspace*{4em}"description": "Set true to draw this shape in white\\
\hspace*{5em}instead of black, which erases along exactly the\\
\hspace*{5em}path the shape would have drawn, at the same stroke\\
\hspace*{5em}width. Defaults to false. The starting shape\\
\hspace*{5em}cannot be erased."\}\},\\
\hspace*{2em}"required": ["polygons"]\}\}\}
\end{quote}
}%

\ok{%
The other tools follow the same form, with the argument name and the field list of Table~\ref{tab:signatures} and an identical \texttt{erase} property.
The tools are declared without strict schema enforcement.
Validation happens in the harness, which skips a malformed operation and appends the skipped count to the tool result (\texttt{OK\ (skipped N malformed)}).
}%

\section{Scorer Validation Detail}
\label{app:scorer-validation}

\ok{%
\autoref{tab:scorervalidation} reports validation statistics for the automated scorers (Section~\ref{sec:scorer-validation}).
}%

\IfFileExists{tables/scorer_validation_note.tex}{
\newcommand{\scorervalidationnote}{ The rating columns are computed over the $2100$ rated drawings. The ViDrA-adapted rating columns are computed over its $423$ held-out test drawings. The human ink column is measured over the $13{,}146$ corpus drawings. The agent ink column is measured over all $2{,}100$ agent drawings.}
}{}
\providecommand{\scorervalidationnote}{}
\begin{table}[htb]
\centering
\caption{%
\ok{%
	Validation of the automated scorers on agent drawings.
	AuDrA is trained on
	the primary subset (11{,}075 drawings) of the AuDrA corpus (13{,}146 rated human drawings)
	ViDrA fits its regression head on the same subset,
	and ViDrA-adapted refits the head on the training split of a 70/10/20 partition of the 2{,}100 rated agent drawings.
	Its rating columns are computed on the held-out test split.
	The first two columns report agreement with the crowdsourced creativity ratings, and the last two columns the Spearman correlation with the inked-pixel baseline on the human and agent corpora.
	Brackets are 95\% confidence intervals.
}%
}%
\label{tab:scorervalidation}
\small
{%
\setlength{\tabcolsep}{3pt}%
\begin{tabular}{lrrrr}
\toprule
Scorer & $r$ with ratings & $\rho$ with ratings & $\rho$ with ink (human) & $\rho$ with ink (agent) \\
\midrule
ViDrA-adapted (ours) & 0.84 [0.81, 0.87] & 0.82 [0.78, 0.86] & 0.72 [0.71, 0.73] & 0.64 [0.61, 0.67] \\
ViDrA (ours) & 0.75 [0.73, 0.77] & 0.72 [0.69, 0.74] & 0.71 [0.70, 0.72] & 0.77 [0.74, 0.78] \\
AuDrA & 0.63 [0.60, 0.66] & 0.61 [0.58, 0.64] & 0.68 [0.67, 0.69] & 0.86 [0.85, 0.87] \\
\bottomrule
\end{tabular}
}%

\end{table}
\FloatBarrier

\section{SVG Ablation}%
\label{app:svg-ablation}%

\ok{%
This appendix reports on the results of comparing the SVG drawing interface with PainterBench's tool-call interface.
}%

\subsection{The SVG Interface}%
\label{app:svg}%

\ok{%
The \textit{interface} manipulation in the sensitivity analysis (Section~\ref{sec:ablations}) withdraws the 
drawing tools of Table~\ref{tab:tools} and exposes one tool in their place, \texttt{draw\_svg}.
This tool takes a string of SVG markup and the same optional \texttt{erase} flag.
The \texttt{undo} and \texttt{finish\_drawing} tools remain available.
The supported subset is \texttt{path} (with its \texttt{M}, \texttt{L}, \texttt{H}, \texttt{V}, \texttt{C}, \texttt{S}, \texttt{Q}, \texttt{T}, \texttt{A}, and \texttt{Z} commands), \texttt{line}, \texttt{polyline}, \texttt{polygon}, \texttt{rect}, \texttt{circle}, and \texttt{ellipse}, nested in optional \texttt{g} groups, with the \texttt{transform} attribute honored.
Markup may be provided as a whole document or a bare list of elements. 
%
The stroke is black (or, when the call erases, white), 
and a \texttt{fill}, \texttt{stroke}, or \texttt{stroke-width} attribute in the markup is ignored.
}%
%
The supported elements and path commands are stated to the agent in the tool description and in the system prompt. An element outside this set is skipped, and the tool result reports the skipped count and the element's tag.

\IfFileExists{figures/contact_sheet_seed2.png}{%
\begin{figure}[htb]
\centering
\includegraphics[width=\linewidth]{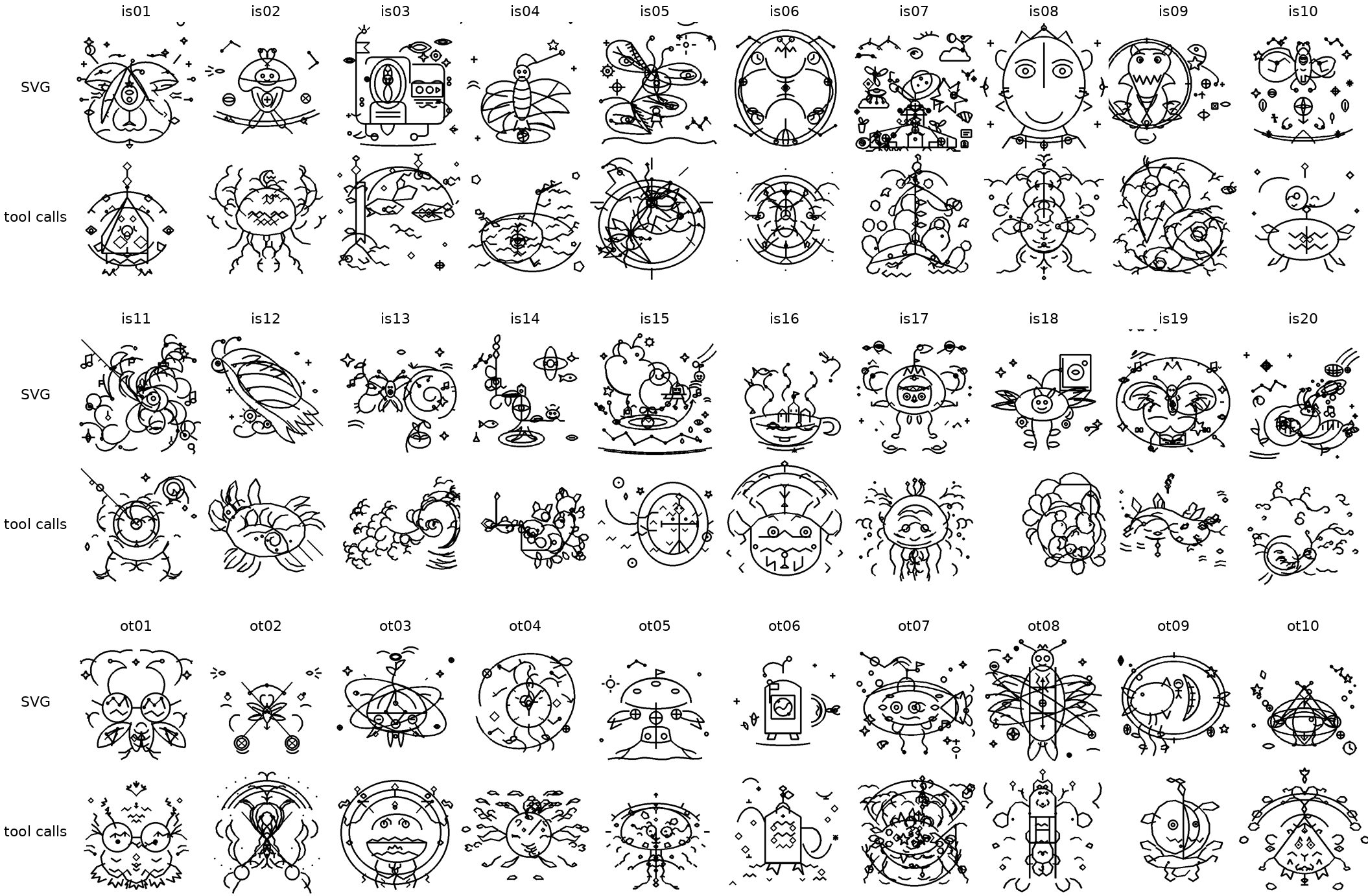}
\caption{Drawings from the SVG interface (top rows) and the paired tool-call trials of the primary study (bottom rows), on the anchor model (GPT-5.6 Luna). 
}
\label{fig:svg-examples}
\end{figure}
}{}

\subsection{Paired comparison of SVG drawings with tool-call drawings}%
\label{app:svgcomparison}%

\label{app:svgresults}%
%
GPT-5.6 Luna is used as the anchor model.
We run 150 trials, five replicates of each of the 30 stimuli, and compare them to the anchor model's drawings from the main study.
Figure~\ref{fig:svg-examples} shows some of the drawings and Table~\ref{tab:svg-interface} compares the condition's process measures with the anchor model's primary-study trials.

\IfFileExists{tables/svg_interface.tex}{%
\begin{table}[htb]
\centering
\caption{%
	Process and rating measures of the interface \rev{manipulation}'s SVG condition and the anchor model's 
	primary-study trials, as per-trial means with standard deviations in parentheses.
}%
\label{tab:svg-interface}
\small
\setlength{\tabcolsep}{4pt}
\resizebox{\linewidth}{!}{%
\begin{tabular}{@{}lrrrrr@{}}
\toprule
 & Rounds & Drawing operations & Operations not drawn & Rated creativity & Rated recognizability \\
\midrule
SVG (n = 150) & 3.3 (1.0) & 57.8 (20.6) & 0.4 (1.1) & 0.68 (0.12) & 0.40 (0.15) \\
Tool calls (n = 150) & 7.5 (1.7) & 70.5 (22.6) & 0.0 (0.2) & 0.61 (0.11) & 0.33 (0.12) \\
\bottomrule
\end{tabular}
}%

\end{table}
}{}

\todo{%
The two interfaces produce similar drawings.
The SVG condition draws in fewer rounds and scores higher on both rated composites, by \ablSvgDelta{} on creativity and \ablSvgRecogDelta{} on recognizability.
These shifts are small against the 0.46 range of the model means (Table~\ref{tab:results}).
Recognizability remains below the human reference.
}%


\textbf{Process.}
\rev{
The SVG trials complete, on average, fewer rounds with fewer drawing operations than the tool-call trials (\autoref{tab:svg-interface}).
The condition issues close to one markup call per drawing round, 342 calls over the 150 trials.
}

\rev{The 150 trials wrote 8{,}676 elements, between 19 and 124 per trial.
Two element types account for 8{,}217 of those, \texttt{path} (5{,}880) and \texttt{circle} (2{,}337), and the rest are \texttt{ellipse} (345), \texttt{line} (84), \texttt{polygon} (16), \texttt{rect} (10), and \texttt{polyline} (4).
Of the 342 strings of markup, 341 parsed as XML,} and no element fell outside the supported subset.
Every SVG interface trial ended by calling the finish tool, and no trial called the \texttt{undo} or \texttt{erase} tools.

\textbf{Content.}
\rev{
The two conditions share content.
The final titles in both draw on a small common vocabulary of cosmic, orbital, clockwork, observatory,
moth, and garden motifs.
The word ``cosmic'' appears in 79 of the 150 SVG titles and 92 of the 150 tool-call titles.
Both conditions center a single subject on the stimulus and surround it with small detached marks, such as stars, crosses, and circles.%
}

\textbf{Ratings.}
\rev{
Table~\ref{tab:svg-interface} reports rated creativity and rated recognizability.
Section~\ref{sec:ablations} reports the rated-creativity contrast.
On the recognizability composite, the SVG condition scores \ablSvgRecogCondMean{} against \ablSvgRecogBaseMean{} for the anchor model's primary-study cells, a paired difference of \ablSvgRecogDelta{} (95\% CI \ablSvgRecogCI, \ablSvgRecogDeltaSd{} SD) over \ablSvgRecogN{} paired stimuli.
The shift matches the rated-creativity shift in size, and the SVG condition's mean recognizability remains below the human reference sample's mean of 0.53 (Table~\ref{tab:results}).
}

\textbf{Execution.}
\rev{
The SVG drawings are composed of smooth closed curves.
The tool-call drawings are composed of short polyline segments with more irregular curvature.
}



\textbf{Failures.}
\rev{%
One failure mode is specific to the SVG interface (cf. ``Operations not drawn'' in \autoref{tab:svg-interface}).
In 48 of the 150 trials, and in 96 elements in total, at least one coordinate was written as an English word rather than as a number, as in }%
\verb|<circle cx="337" cy=" ninety" r="10"/>|.
\rev{%
Fifty-seven of the 96 elements were left with no geometry the renderer could use. Each of these elements was reported back to the agent as not drawn.
In the remaining 39 elements, the affected attribute took its SVG default of zero and the element was drawn at the edge of the canvas.
PainterBench's typed tools do not admit the second outcome, because a coordinate there is a schema-typed integer and a value that does not parse is counted as a malformed operation and skipped.
}%

\section{Process Markers}
\label{app:processproduct}

\noindent
\ok{%
This appendix defines the \ok{eleven} process markers of Section~\ref{sec:eval} and reports their correlation to rated creativity (i.e., crowdsourced creativity ratings).
Each marker is computed from the tool-call trace of a trial.
%
%
Table~\ref{tab:markerdefs} defines each marker and names the related MTCI measure \citep{barbot2018mtci}. 
The five markers of the second block have no MTCI counterpart.
}%

\begin{table}[!htb]%
\centering
\caption{%
\ok{%
	Process markers computed from the tool-call trace of a trial run.
	Rows marked $\ast$ are the four components included in the process effort index (Section \ref{sec:process}).
}%
}%
\label{tab:markerdefs}%
\small
\begin{tabular}{@{}llp{6.5cm}@{}}
\toprule
Marker & Substitutes for MTCI & Definition \\
\midrule
Exploration rounds          & Exploration phase & Rounds before the round of the first ink operation \\
Tool calls $\ast$           & Production phase  & Ink calls from the first to the last ink operation \\
Drawing operations $\ast$   & Production phase  & Operations contained in those calls \\
Verification rounds         & Verification phase & Rounds after the last inking round that contain no ink call and do not finish the trial \\
Rounds to completion $\ast$ & Response time     & Rounds in the trial \\
Tool diversity $\ast$       & Flexibility       & Shannon entropy of trial's tool-type distribution \\
\midrule
Drawing operations per round & ---              & Ink operations divided by rounds \\
Parallel rounds             & ---               & Rounds carrying more than one tool call \\
Largest round               & ---               & Tool calls in the round that carried the most \\
Undo calls (\%)             & ---               & Calls to the revert tool, as a percentage of the trial's tool calls \\
Erase calls (\%)            & ---               & Calls that set the \texttt{erase} flag, as a percentage of the trial's tool calls \\
\bottomrule
\end{tabular}
\end{table}
\FloatBarrier

\ok{%
Table~\ref{tab:processproduct} reports two Spearman correlations per marker, each with a 95\% confidence interval.
Exploration rounds are zero in every trial, so no correlation is reported for this marker.
}%
\todo{%
The pooled correlation is computed over all rated drawings. 
The within-model correlation is the mean of the per-model correlations, and its confidence interval resamples drawings within each model.
}%

\IfFileExists{tables/process_product_note.tex}{
\newcommand{\processproductnote}{ The correlations are computed over the $2100$ rated drawings of the $14$ models.}
}{\newcommand{\processproductnote}{ \todo{Values are placeholders pending data collection.}}}
\begin{table}[htb]%
\centering
\caption{%
\ok{%
	Spearman correlation between each process marker and rated creativity, pooled over all rated drawings and as the mean of the per-model correlations.
}%
}%
\label{tab:processproduct}%
\small
\IfFileExists{tables/process_product.tex}{
{%
\setlength{\tabcolsep}{3pt}%
\begin{tabular}{lrrrr}
\toprule
& \multicolumn{2}{c}{Pooled} & \multicolumn{2}{c}{Within-model} \\
\cmidrule(lr){2-3}\cmidrule(lr){4-5}
Marker & $\rho$ & 95\% CI & $\bar{\rho}$ & 95\% CI \\
\midrule
Tool calls & 0.06 & [0.01, 0.11] & -0.01 & [-0.05, 0.04] \\
Drawing operations & 0.53 & [0.50, 0.56] & 0.06 & [0.02, 0.11] \\
Verification rounds & -0.05 & [-0.11, 0.01] & -0.06 & [-0.11, 0.02] \\
Rounds to completion & 0.12 & [0.08, 0.17] & -0.03 & [-0.08, 0.01] \\
Tool diversity & 0.33 & [0.29, 0.37] & 0.13 & [0.09, 0.17] \\
\midrule
Drawing operations per round & 0.59 & [0.56, 0.62] & 0.16 & [0.12, 0.20] \\
Parallel rounds & -0.26 & [-0.30, -0.23] & 0.06 & [0.00, 0.12] \\
Largest round & -0.27 & [-0.30, -0.23] & 0.07 & [0.01, 0.13] \\
Undo calls (\%) & -0.01 & [-0.06, 0.03] & -0.08 & [-0.13, -0.02] \\
Erase calls (\%) & -0.01 & [-0.05, 0.03] & -0.05 & [-0.09, 0.00] \\
\bottomrule
\end{tabular}
}%
}{%
\begin{tabular}{lrrrr}%
	\toprule
	& \multicolumn{2}{c}{Pooled} & \multicolumn{2}{c}{Within-model} \\
	\cmidrule(lr){2-3}\cmidrule(lr){4-5}
	Marker & $\rho$ & 95\% CI & $\bar{\rho}$ & 95\% CI \\
	\midrule
	Tool calls      & -- & -- & -- & -- \\
	Drawing operations & -- & -- & -- & -- \\
	Verification rounds   & -- & -- & -- & -- \\
	Rounds to completion  & -- & -- & -- & -- \\
	Tool diversity        & -- & -- & -- & -- \\
	\midrule
	Drawing operations per round & -- & -- & -- & -- \\
	Parallel rounds       & -- & -- & -- & -- \\
	Largest round         & -- & -- & -- & -- \\
	Undo calls (\%)       & -- & -- & -- & -- \\
	Erase calls (\%)      & -- & -- & -- & -- \\
	\bottomrule
\end{tabular}
}
\end{table}
\FloatBarrier

%

\section{Crowdsourced Ratings}
\label{app:crowdosurcing}

\subsection{Rater Instruction}
\label{app:raterinstruction}

\ok{%
The following instruction was shown in the crowdsourcing task 
before the worker started the work:%
%
%
\begin{quote}\footnotesize\ttfamily
In this task you will rate a series of black-and-white line drawings. Each drawing started from an incomplete shape, which the artist completed into a full drawing.
\\
\\
For each drawing you will answer two questions.
\\
\\
1. How creative is this drawing?
Rate from 1 (``Not At All Creative'') to 5 (``Very Creative''). Focus on how creative the idea is, not how artistic or skillfully drawn it is.
\\
\\
2. Does this drawing show a recognizable object or scene?
Rate from 1 (``Not At All Recognizable'') to 5 (``Very Recognizable''). A drawing is recognizable if you can tell what it depicts.
\\
\\
There are no right or wrong answers. Rate each drawing on its own, and use the full range of the scale when the drawings differ. Some drawings may \rev{be} difficult to rate, but please make an honest effort to rate each one.
\\
\\
Do not use your browser's back button. After the last drawing, click Submit to finish.
\end{quote}
}%

\subsection{Rater Allocation}
\label{app:power}

\ok{%
Each drawing is rated on both creativity and recognizability by $k$ raters.
This appendix describes how we select $k$.
}%

\ok{%
Let $\rho_1$ denote the expected correlation between two ratings of the same drawing made by different raters.
This is the one-way intraclass correlation, in which everything that varies between the two ratings counts as error, including rater severity, a rater's systematic tendency to rate low or high.
By the Spearman--Brown formula~\citep{spearman1910faulty,brown1910experimental}, the average of $k$ such ratings (i.e., the composite) has reliability $\rho_k = k\rho_1 / (1 + (k-1)\rho_1)$.
}%

\ok{%
%
Our design target is a composite whose reliability matches that of the human ratings.
In the released individual ratings of the AuDrA primary corpus, a pool of 50 raters rated 11{,}075 drawings with a median of 8~ratings per drawing, which gives $\rho_1 = .39$ on rated creativity and, by Spearman--Brown, a composite reliability of $.84$ at the median count~\citep{patterson2024audra}.
As the minimum, we set the target $\rho_k \geq .75$, the threshold for good reliability in the guidelines of \citet{koo2016guideline}.
Solving $\rho_k \geq .75$ for the number of raters gives the smallest count that reaches the target, $k = \lceil 3(1-\rho_1)/\rho_1 \rceil$.
We call this formula the \textit{allocation rule}.
}%

\ok{%
The single-rater reliability $\rho_1$ is a property of the rater pool and the rating questions, and it is unknown before data collection.
We therefore estimate $\rho_1$ in a pilot and fix $k$ before the data collection.
The pilot runs the full rating protocol on one batch of drawings ($N=30$) and yields one estimate of $\rho_1$ per question.
To guarantee the target under the sampling error of this single batch, we apply the allocation rule for each question at the one-sided 95\% lower confidence bound of its estimate, computed by a bootstrap over drawings.
We collect the larger of the two resulting counts, up to a cap of 25 ratings per drawing.
}%

\begin{table}[htb]
\centering
\caption{The number of ratings per drawing, $k$, required to reach the reliability target $\rho_k \geq .75$ at each single-rater reliability $\rho_1$. The last column is the total ratings per question for the \rev{$3{,}000$} rated drawings. The bold row, the lower confidence bound of the pilot's creativity estimate, sets the collected count.}
\label{tab:power}
\small
{%
\setlength{\tabcolsep}{3pt}%
\begin{tabular}{rrrrrr}
\toprule
Single-rater & $k$ & $k$ & Reliability & Attenuation & Ratings \\
ICC $\rho_1$ & demanded & collected & $\rho_k$ & $\sqrt{\rho_k}$ & total \\
\midrule
.05 & 57 & 25 & .568 & .754 & 75{,}000 \\
.10 & 27 & 25 & .735 & .857 & 75{,}000 \\
.15 & 17 & 17 & .750 & .866 & 51{,}000 \\
.20 & 12 & 12 & .750 & .866 & 36{,}000 \\
\textbf{.21} & \textbf{12} & \textbf{12} & \textbf{.762} & \textbf{.873} & \textbf{36{,}000} \\
.25 & 9 & 9 & .750 & .866 & 27{,}000 \\
.30 & 7 & 7 & .750 & .866 & 21{,}000 \\
.33 & 7 & 7 & .773 & .879 & 21{,}000 \\
\bottomrule
\end{tabular}
}%

\end{table}
\FloatBarrier

\ok{%
Table~\ref{tab:power} tabulates the allocation rule over a range of $\rho_1$ values.
The pilot 
estimated $\rho_1 = .33$ for creativity and $\rho_1 = .42$ for recognizability (Krippendorff's ordinal $\alpha = .31$ and $.41$, respectively), with lower confidence bounds of $.21$ and $.29$.
At these bounds, the allocation rule demands $k = 12$ for creativity and $k = 8$ for recognizability, so we collect $k = 12$ ratings per question.
At 
this number of raters, the creativity composite has reliability $.85$ and the recognizability composite $.90$ at the point estimates, and $.76$ and $.83$ at the lower bounds.
\todo{At the point estimate, the creativity composite reaches the composite reliability of the AuDrA human ratings.}
}%

\section{Title Semantics}%
\label{app:titles}%

\ok{%
The AuDrA corpus records participant-written titles 
for two of its four sets 
(1{,}349 drawings).
%
%
The title is a channel on which the two populations (humans and AI) can be compared, and it states what the drawing was meant to be.
We report three measures over the 
human-generated and AI-generated titles.
The \emph{abstraction-marker rate} is the share of titles containing a term from a fixed list of twenty-six words that name a configuration rather than a thing (e.g., \emph{abstract}, \emph{geometric}, \emph{composition}, \emph{pattern}, etc.).
\emph{Concreteness} is the mean concreteness of a title's content words under 
the word-concreteness ratings of
\citet{brysbaert2014concreteness}.
\emph{Separability} is the \rev{stratified five-fold} cross-validated area under the ROC curve of a logistic regression that predicts the population from the title's sentence embedding.
Titles are embedded with the all-mpnet-base-v2 sentence-transformer model \citep{reimers2019sentencebert}.
Figure~\ref{fig:title-examples} depicts example drawings and their agent-assigned titles, showing clear differences in agent-generated titles. 
}%

\IfFileExists{figures/examples_grid.png}{%
\begin{figure}[hbt]
\centering
\includegraphics[width=\linewidth]{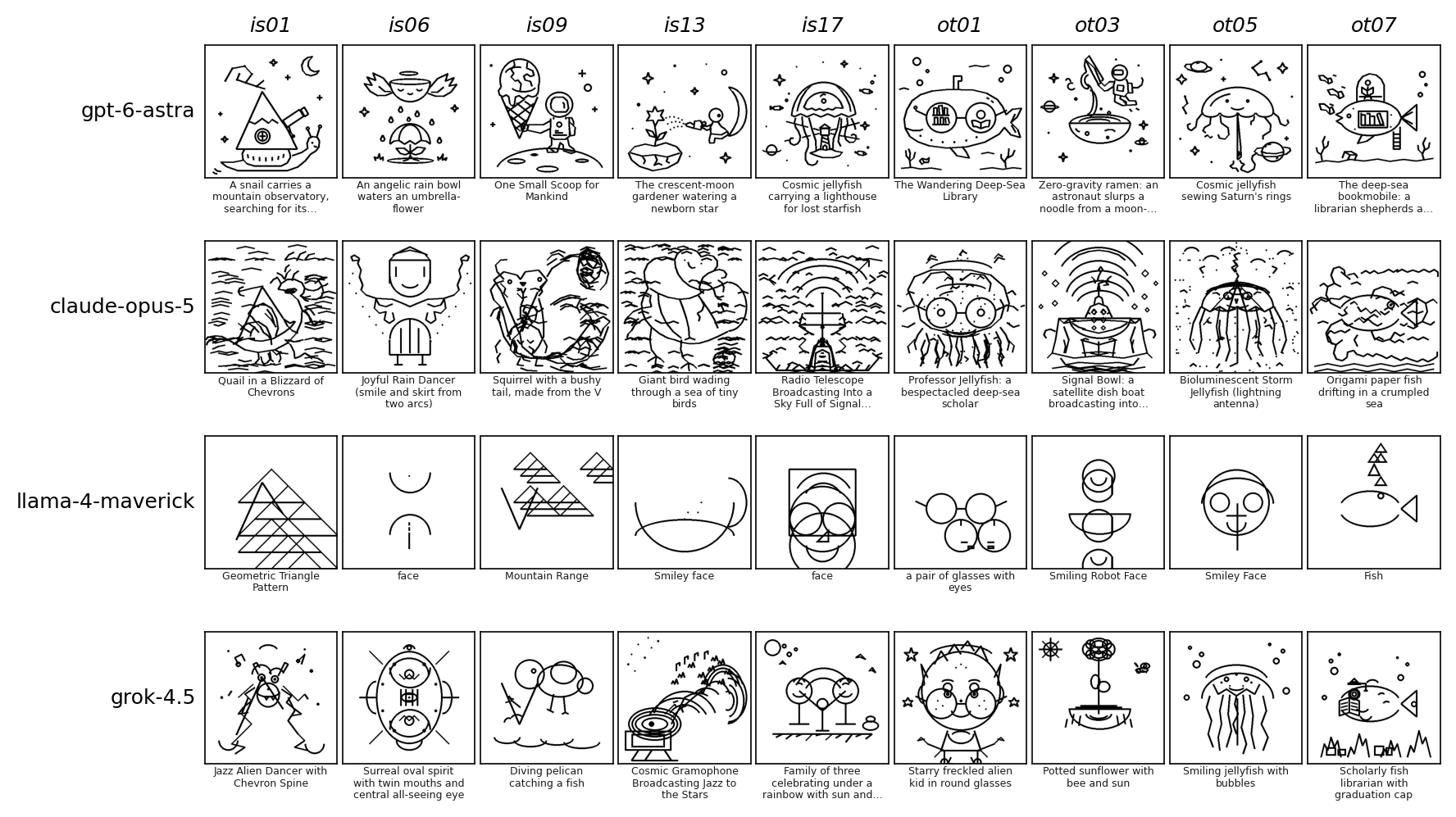}
\caption{Examples of agent-chosen titles.}
\label{fig:title-examples}
\end{figure}
\FloatBarrier
}{}%

\newcommand{\titlenagent}{2,099}
\newcommand{\titlenuntitled}{1}
\newcommand{\titlenhuman}{1,349}
\newcommand{\titlenmodels}{14}
\newcommand{\titleabstractionagent}{10.58\%}
\newcommand{\titleabstractionhuman}{0.67\%}
\newcommand{\titleabstractiondiff}{+9.91 points, 95\% CI [8.62, 11.46]}
\newcommand{\titleconcretenessagent}{4.04}
\newcommand{\titleconcretenesshuman}{4.56}
\newcommand{\titleconcretenessdiff}{-0.52, 95\% CI [-0.56, -0.48], $d = -0.98$}
\newcommand{\titleseparability}{0.97}
\newcommand{\titleseparabilitylength}{0.88}
\newcommand{\titlelengthagent}{6.11}
\newcommand{\titlelengthhuman}{2.32}
\newcommand{\titlelengthdiff}{+3.78, 95\% CI [3.60, 3.94], $d = 1.37$}
\ok{%
The human and agent titles differ on all three measures.
Abstraction markers appear in \titleabstractionagent{} of the 
agent titles, but only in \titleabstractionhuman{} of the 
human titles in AuDrA's corpus. 
Mean concreteness is \titleconcretenessagent{} for agent titles and \titleconcretenesshuman{} for human titles on the five-point scale of \citet{brysbaert2014concreteness} (\rev{difference }\titleconcretenessdiff).
A logistic regression on the sentence embeddings predicts the population with a cross-validated AUC of \titleseparability{}.
Mean title length is \titlelengthagent{} words for agent titles and \titlelengthhuman{} words for human titles (difference \titlelengthdiff).
A classifier given the word count alone reaches an AUC of \titleseparabilitylength{}, so part of the separability is carried by title length.
}%

\IfFileExists{figures/title_umap.pdf}{%
\begin{figure}[hbt]
\centering
\includegraphics[width=0.75\linewidth]{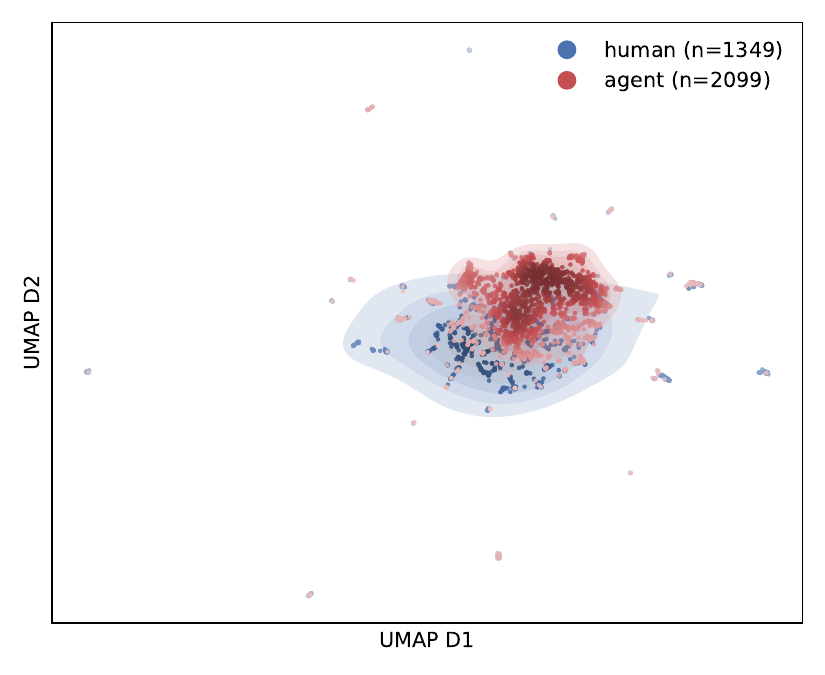}
\caption{Two-dimensional UMAP projection\rev{~\citep{mcinnes2018umap}} of sentence embeddings \rev{of the agent titles and the participant titles of the AuDrA corpus}.
	\ok{%
		Each population is shaded by its kernel density estimate.
	}%
}
\label{fig:title-umap}
\end{figure}
}{%
\begin{figure}[htb]
\centering
\fbox{\rule{0pt}{0.32\linewidth}\rule{0.6\linewidth}{0pt}}
\caption{\todo{[Placeholder] Two-dimensional projection of sentence embeddings of the titles agents gave their drawings and the titles participants gave theirs in the AuDrA corpus.
}}
\label{fig:title-umap}
\end{figure}
}

\FloatBarrier

\ok{%
Figure~\ref{fig:title-umap} projects the sentence embeddings of the agent and human titles into two dimensions.
%
The agent titles concentrate in a narrower region of the embedding space than the human titles, consistent with the abstraction-marker and concreteness differences above.
Table~\ref{tab:title-tfidf} lists the ten highest-weighted TF-IDF terms of each population.
While humans name persons, animals, and everyday objects, agents often depict robots or mix object nouns with the configuration terms \emph{abstract} and \emph{geometric}.
The highest-weighted agent term, \emph{cosmic}, is a modifier shared across 12 of the \titlenmodels{} models.%
}%

\IfFileExists{tables/title_tfidf.tex}{%
\begin{table}[htb]
\centering
\caption{The ten highest-weighted TF-IDF terms in the agent titles and in the participant-written titles of the AuDrA corpus. \emph{Share} is the share of that population's titles containing the term.}
\label{tab:title-tfidf}
\small
{%
\setlength{\tabcolsep}{3pt}%
\begin{tabular}{l r p{0.28\linewidth} l r p{0.20\linewidth}}
\toprule
Agent term & Share & Example title & Human term & Share & Example title \\
\midrule
cosmic & 10.0\% & \emph{Cosmic Clockwork Fish} & house & 6.3\% & \emph{a house} \\
face & 6.9\% & \emph{a face} & face & 6.5\% & \emph{a face} \\
robot & 8.1\% & \emph{Robot Face} & bird & 4.0\% & \emph{a bird} \\
abstract & 5.4\% & \emph{Abstract Composition} & person & 3.4\% & \emph{a person} \\
sea & 4.8\% & \emph{Lighthouse by the sea} & boat & 2.7\% & \emph{a boat} \\
geometric & 4.2\% & \emph{Geometric Composition} & man & 3.0\% & \emph{a man} \\
moon & 5.0\% & \emph{Constellation with Moon} & tree & 2.2\% & \emph{a tree} \\
fish & 3.6\% & \emph{Fish} & pizza & 1.8\% & \emph{a pizza} \\
jellyfish & 3.5\% & \emph{Jellyfish} & arrow & 1.6\% & \emph{an arrow} \\
creature & 4.2\% & \emph{Whimsical creature face} & pencil & 1.5\% & \emph{a pencil} \\
\bottomrule
\end{tabular}
}%

\end{table}
}{}


\end{document}